\documentclass{article}
\usepackage{ijcai25}

\usepackage{times}
\usepackage{soul}
\usepackage{url}
\usepackage[hidelinks]{hyperref}
\usepackage[utf8]{inputenc}
\usepackage[small]{caption}
\usepackage{graphicx}
\usepackage{amsmath}
\usepackage{amsthm}
\usepackage{booktabs}
\usepackage{algorithm}
\usepackage{algorithmic}
\usepackage[switch]{lineno}

\usepackage{amssymb,amsfonts,amsmath}
\usepackage{textcomp}
\usepackage{xcolor}
\usepackage{multicol}
\usepackage{multirow}
\usepackage{makecell}
\usepackage{colortbl}
\usepackage{color}
\usepackage{subcaption}
\usepackage{boldline}
\usepackage{array}
\usepackage{mathrsfs}
\usepackage{float}

\title{\model: A Model to Disentangle Aperiodic Events from Traffic Series}

\author{
Xinyu Su\and
Feng Liu\and
Yanchuan Chang\and
Egemen Tanin\and
Majid Sarvi\And
Jianzhong Qi\footnote{Corresponding author.}
\\
\affiliations
The University of Melbourne\\
\emails
\{suxs3@student., feng.liu1@, yanchuan.chang@, etanin@, majid.sarvi@, jianzhong.qi@\}unimelb.edu.au
}

\newcommand{\model}{DualCast}
\newcommand{\argmin}{\mathop{\mathrm{arg\,min}}}

\begin{document}

\maketitle

\begin{abstract}
Traffic forecasting is crucial for transportation systems optimisation. 
Current models minimise the \emph{mean} forecasting errors, often favouring periodic events prevalent in the training data, while \emph{overlooking critical aperiodic ones} like traffic incidents.
To address this, we propose \emph{\model}, a dual-branch framework that disentangles traffic signals into intrinsic {spatial-temporal} patterns and external environmental contexts, including aperiodic events.
\model\ also employs a cross-time attention mechanism to capture high-order spatial-temporal relationships from both periodic and aperiodic patterns. \model\ is versatile. We integrate it with recent traffic forecasting models, consistently reducing their forecasting errors by up to 9.6\% on multiple real datasets. Our source code is available at \url{https://github.com/suzy0223/DualCast}.
\end{abstract}

\section{Introduction}

Traffic forecasting is essential for intelligent transportation systems (ITS), enabling real-time solutions like route planning and transportation scheduling.

Deep learning-based solutions have dominated the traffic forecasting literature in recent years. 
They typically adopt graph neural networks (GNNs) for modelling spatial patterns and sequential models for modelling temporal patterns~\cite{stsgcn,stfgnn,gcnode,gamcn,stgnn,liu2022contrastive}.
Besides, a series of recent studies adopt the attention mechanism to capture dynamic relationships in traffic patterns~\cite{astgcn,staeformer,predformer}.

These solutions are primarily designed to minimise \emph{mean} forecasting errors, a common evaluation metric~\cite{sttn,gman,pdformer}. This optimisation focuses on \emph{periodic} traffic patterns, which are both easier to forecast and more prevalent in the traffic data, resulting in an easier reduction in mean errors.

These models struggle with rare and random \emph{aperiodic} events, such as traffic incidents, making them difficult to forecast.
However, promptly identifying and adapting to such events is essential for effective real-time traffic forecasting.

\begin{figure}[t]
         \centering
  \includegraphics[width=0.49\textwidth]{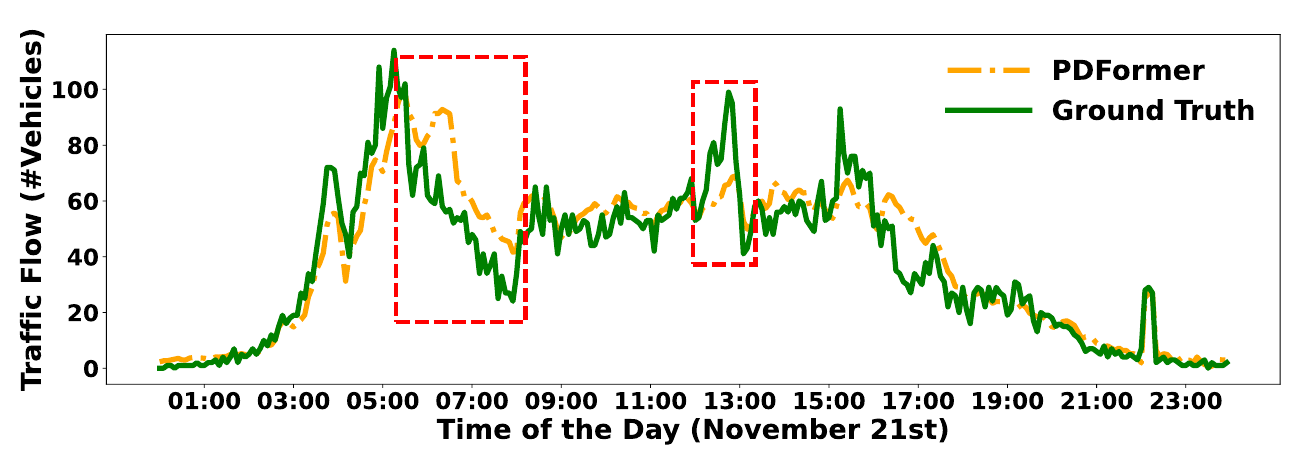}
         \caption{
         An example of a recent model PDFormer forecasting 60-minute-ahead traffic flows. The model has strong overall results but fails to respond to sudden changes (highlighted by the red boxes).
         }
         \label{fig:motivation}
\end{figure}

Fig.~\ref{fig:motivation} shows PDFormer~\cite{pdformer} forecasting the traffic flow on a California freeway 60 minutes ahead for one day (detailed in Section~\ref{subsec:exp_results}). There are two substantial gaps between the forecasts (the orange dashed line) and ground truth (the green solid line) at around 07:00 and 13:00. These gaps would be overlooked if only mean forecasting error is considered, as the overall patterns are similar. 

In this paper, we propose \emph{\model} -- a model framework to address the issue above. \model\ is \emph{not} yet another traffic forecasting model. Instead, we aim to present \textbf{a generic structure to power current traffic forecasting models with stronger learning capability to handle aperiodic patterns from traffic series.} 
\model\ has a dual-branch design to disentangle a traffic observation into two signals: (1)~the \emph{intrinsic branch} learns intrinsic (periodic) \emph{spatial-temporal patterns}, 
and (2) the \emph{environment branch} learns  external environment contexts that contain aperiodic patterns.
We implement \model\ with three representative traffic forecasting models, that is, STTN~\cite{sttn}, GMAN~\cite{gman} and PDFormer~\cite{pdformer}, due to their reported strong learning outcomes.

The success of our dual-branch framework relies on three loss functions: \emph{filter loss}, \emph{environment loss}, and \emph{DBI loss}. These functions guide \model\ to disentangle the two types of signals and later fuse the learning outcomes to generate the forecasting results. 

(1)~The \textbf{filter loss} computes the reciprocal of Kullback-Leibler (KL) divergence between the feature representations learned from two branches, ensuring that each branch captures distinct signals from the input.
(2)~The \textbf{environment loss} is designed for the environment branch. It computes the reciprocal of KL divergence between a batch of training samples and a randomly permuted sequence of those samples in the same batch. This loss encourages \model\ to learn \emph{the diverse environment contexts} at different times, as the samples of the training pair used in the KL divergence are drawn from different periods. 
(3)~The \textbf{DBI loss} is designed for the intrinsic branch. It encourages \model\ to learn more separated representations for training samples with different (periodic) traffic patterns while closer representations for samples within the same traffic patterns.  

 The three models~\cite{gman,sttn,pdformer} with which \model\ is implemented all use self-attention. To enhance \textbf{their self-attention modules to capture spatial-temporal correlations}, we identify two issues:
(1)~Existing self-attention-based models~\cite{pdformer,gman,sttn}  
learn spatial and temporal patterns separately, focusing on nodes at the same time step or the same node across time. They neglect correlations between different nodes across time, while such correlations are important for modelling the impact of aperiodic events like the impact of traffic incidents propagating spatially over time. 
(2)~Existing models take either a local attention~\cite{pdformer} or a global attention~\cite{gman} setup. They compute attention only among connected nodes (based on the adjacency matrix) or among all nodes. This limits receptive fields or loses hierarchical relationships critical for traffic flow propagation. 

To address these issues, we propose:
(1)~a \emph{cross-time attention} module using hierarchical message passing based on a conceptual space-time tree, enabling attention across nodes and time steps to better model spatial-temporal traffic propagation without extra storage or computational overhead.
(2)~an \emph{attention fusion} module to combine local and global attention, expanding the receptive field and capturing hierarchical node relationships.

Overall, this paper makes the following contributions:

(1)~We propose \model\ -- a model framework equipped with two branches and three loss functions to disentangle complex traffic observations into two types of signals for more accurate forecasting. \model\ is versatile in terms of the models to form its two branches -- we use self-attention-based models for their reported strong learning outcomes. 

(2)~We propose two enhancements for self-attention-based forecasting models: (i)~A cross-time attention module to capture high-order spatial-temporal correlations, and (ii)~An attention fusion module to combine global and local attention, enlarging \model's receptive field and learning the hierarchical relationships among the nodes. 
 
(3)~We conduct experiments on both freeway and urban traffic datasets, integrating \model\ with three self-attention-based models GMAN, STTN, and PDFormer. The results show that:~(i)~\model\ consistently reduces the forecasting errors for these models, with stronger improvements at times with more complex environment contexts and by up to 9.6\% in terms of RMSE;~(ii)~\model\ also outperforms the SOTA model consistently and by up to 2.6\%.

\section{Related Work}
\label{sec:related_work}
Traffic forecasting typically employs sequence models~\cite{arima,lstm,graphwavenet} to capture temporal patterns and GNNs for spatial correlations~\cite{tian2015predicting,arima_traffic,lstm_traffic,stgcn,jin2022selective,increase,ma2024learning,STSM}. Spatial and temporal layers can be arranged sequentially or in parallel, and fused via methods such as gated fusion~\cite{gatedfusion}. Some GNN-based models~\cite{hyTemporalGraph} connect graph snapshots over time to reduce the negative impact of the ripple effect, which still overlooks time-varying relationships. Self-attention based traffic forecasting models handle this issue easily~\cite{staeformer,stabc}. 

Moreover, some studies disentangle traffic series into periodic components~\cite{chen2021traffic,stnorm,stwave,museNet,modwavemlp,fouriergnn}, different levels~\cite{TimeDRL}, or invariant and environment signals~\cite{cast,zhou2023maintaining}. Others adopt memory augmentation to enhance sensitivity to aperiodic signals~\cite{eastnet,megaCRN}. Our proposed \model\ employs a dual-branch structure with three loss functions and cross-time attention to flexibly capture diverse, aperiodic patterns and environment contexts without relying on predefined patterns. A full discussion is provided in Appendix~\ref{sec:a_related_work}.

\section{Preliminaries}
\label{sec:Preliminaries} 
\paragraph{Traffic forecasting.} We model a network of traffic sensors as a graph $G=(V, E, \mathbf{A})$, where $V$ denotes a set of $N$ nodes (each representing a sensor) and $E$ denotes a set of edges representing the spatial connectivity between the sensors based on the underlying road network. $\mathbf{A}\in \mathbb{R}^{N\times N}$ is an adjacency matrix derived from the graph. If $v_{i},v_{j} \in V$ and $(v_i,v_j) \in E$, then $\mathbf{A}_{i,j}=1$; otherwise, $\mathbf{A}_{i,j}=0$. 

For each sensor (i.e., a \emph{node} hereafter, for consistency) $v_i \in V$, we use $x_{i,t} \in \mathbb{R}^C$ to represent the traffic observation of $v_i$ at time step $t$, where $C$ is the number of types of observations, e.g., traffic flow and traffic speed. 
Further, $\mathbf{X}_t=[x_{1,t}, x_{2,t}\ldots,x_{N,t}] \in \mathbb{R}^{N \times C}$ denotes the observations of all nodes in $G$ at time step $t$, while $\hat{\mathbf{X}}_t \in \mathbb{R}^{N \times C}$ denotes the forecasts of the nodes in $G$ at time step $t$.  We use $\mathbf{X}_{t_i:t_j}$ to denote the consecutive observations from $t_i$ to $t_j$.

\begin{figure*}[t!]
         \centering
  \includegraphics[width=0.9\linewidth]{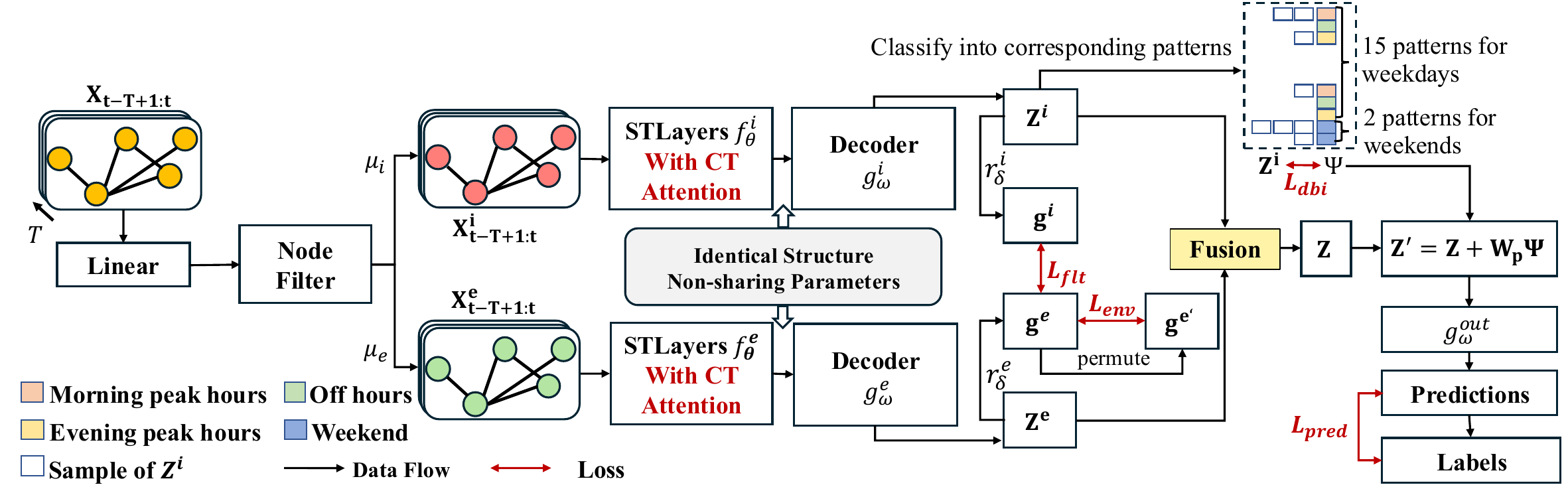}
         \caption{The \model\ framework. \model\ maps the input traffic observations $\mathbf{X}_{t-T+1:t}$ ($\mathbf{X}$, for simplicity) into a $D$-dimensional space and uses a node filter to disentangle them into intrinsic signals ($\mathbf{X}^i$) and environment signals ($\mathbf{X}^e$). 
         Each signal is fed into a separate branch (intrinsic or environment branch) formed by an encoder (STLayers with CT attention), a decoder, and a function generating traffic representation  $\mathbf{Z^i}$ ($\mathbf{Z^e}$) and graph representation $\mathbf{g}^i$ ($\mathbf{g}^e$). 
         Three loss functions are designed to optimise \model: (1)~ Filter loss computes KL divergence between $\mathbf{g}^i$ and $\mathbf{g}^e$ to guide each branch to capture distinct signals from the input. (2)~The environment loss computes KL divergence between $\mathbf{g}^e$ and  $permute(\mathbf{g}^e)$ to encourage \model\ to learn different environment contexts for different times, and (3)~The DBI loss promotes learning distinctive representations for different periodic traffic patterns.  
         \model\ finally fuses $\mathbf{Z^i}$, $\mathbf{Z^e}$, and $\Psi$ to produce $\mathbf{Z}'$, which is then mapped into the output space and compared with the ground truth to compute the prediction loss.}
         \label{fig:framework}
\end{figure*}

\paragraph{Problem statement.} Given a sensor graph $G = (V, E, \textbf{A})$, a traffic forecasting model generally adopts an encoder-decoder structure to learn from the traffic observations of the previous $T$ steps and generate forecasts for the following $T'$ steps $\hat{\mathbf{X}}_{t+1\,:\,t+T'} = g_\omega(f_\theta(\mathbf{X}_{t-T+1\,:\,t}))$,
where $f_\theta$ and $g_\omega$ denote the encoder and the decoder, respectively, and $\theta\in\Theta$ and $\omega\in\Omega$ denote the learnable parameters.
We aim to find $f_\theta$ and $g_\omega$ to minimise the errors between the ground-truth  observations and the forecasts:
{\small
\begin{equation}
\label{eq:pred_loss}
\argmin_{\theta\in\Theta, \; \omega\in\Omega}\mathop{\mathbb{E}}_{t\in\mathcal{T}}
\big|\big|\,
g_\omega \big(f_\theta(\mathbf{X}_{t-T+1\,:\,t})\big) - \mathbf{X}_{t+1\,:\,t+T'}
\,\big|\big|_{p}, 
\end{equation}}
where $\mathcal{T}$ denotes the time range of traffic observations of the dataset, and $p$ is commonly set as 1 or 2. 

We propose a model optimisation framework named \model\ compatible with recent self-attention-based (\emph{STF} hereafter) models~\cite{pdformer,gman,sttn,STPGNN}. Due to space limit, we detail these models in Appendix~\ref{sec:a_method}.

\section{The \model\ Framework}
Fig.~\ref{fig:framework} shows our proposed \model\ with a \emph{dual branch} structure (Section~\ref{subsec:dual_branches}), which disentangles traffic observations into two types of underlying signals, namely \emph{intrinsic (periodic) signals} and \emph{environmental (aperiodic) signals} for accurate traffic forecasting. We introduce three loss functions: \emph{filter loss}, \emph{environment loss}, and \emph{DBI loss}, to guide the model to generate distinct representations for these signals. 

As self-attention-based traffic forecasting models have competitive performance, we use them as baseline models. We design \emph{rooted sub-tree cross-time attention} (CT attention) module (Section~\ref{subsec:ct-attention}) which can efficiently capture dynamic and high-order spatial-temporal correlations between sensors on both branches to enhance their performance.

\subsection{Dual-branch Structure and Optimisation}
\label{subsec:dual_branches}
\textbf{Dual-branch structure.}
The dual-branch structure disentangles the traffic observations into intrinsic and environment signals. The intrinsic branch (IBranch) and environment branch (EBranch) share an identical structure with separate parameters.
The intrinsic signals reflect intrinsic (periodic) traffic patterns, while the environment signals reflect external environment (aperiodic) contexts, such as traffic incidents. The two signals together determine the traffic forecasts.

Given a batch of input observations $\textbf{X} \in \mathbb{R}^{B \times T \times N \times C}$, we compute \emph{disentangling coefficients} $\mu_i, \mu_e$ for intrinsic and environment signals as $\mu_i, \mu_e = {\rm softmax}({\rm Linear}(\mathbf{X}))$,
where {\rm Linear} denotes a linear layer with an output size of 2; 
Both $\mu_i$ and $\mu_e$ have shape $\mathbb{R}^{B\times T\times N}$. 
Then, we produce the intrinsic signals $\mathbf{X}^{i}=\mu_i\odot\mathbf{X}$ and the environment signals $\mathbf{X}^{e}=\mu_e\odot\mathbf{X}$, where $\odot$ is element-wise product, and $\mu_i$, $\mu_e$ are expanded along the last dimension.
These signals are fed into IBranch and EBranch to generate representations $\mathbf{Z^i}$ and $\mathbf{Z^e}$, respectively. 
The process is detailed for IBranch, with EBranch operating similarly.

In IBranch, $\mathbf{X}^{i}$ is fed into the spatial-temporal encoder $f_{\theta}^{i}$ to produce a hidden representation $\mathbf{H}^{i}$, which is passed to the decoder $g_{\omega}^{i}$ to produce the output representation of the branch, $\mathbf{Z^i} \in R^{B\times T\times N \times D}$. We concatenate the outputs of both branches to obtain $\mathbf{Z} = \rm{Concat}(\mathbf{Z^i}, \mathbf{Z^e})$ and  generate the forecasts $\hat{\mathbf{X}}$ from $\mathbf{Z}$ through another linear layer $g_{\omega}^{out}$.

\paragraph{Model training.} We train \model\ using three loss functions: (1)~filter loss to separate branch feature spaces, (2)~environment loss to learn the impact of environment contexts, and (3)~DBI loss to learn different periodic patterns.

\paragraph{Filter loss.} Denoted as $L_{flt}$, is based on KL divergence. It encourages each branch to capture distinct signals.
We aggregate the output $\mathbf{Z}^{i}$ from IBranch along the time dimension by a linear layer and then along node with mean pooling to produce an overall representation $\mathbf{g}^i \in \mathbb{R}^{B\times D}$ of each input sample. This process is denoted as $r_\delta^i$, where $\delta$ refers to linear layer  parameters.
Similarly, we obtain $\mathbf{g}^e$ through $r_\delta^e$ from EBranch.
We use ${\rm softmax(\cdot)}$ to map $\mathbf{g}^e$ and $\mathbf{g}^i$ into distribution and compute the filter loss as follows:
{\small
\begin{equation}
\label{eq:filter_loss}
L_{flt} = {\rm{KL}\big(\mathbf{g}^i,\mathbf{g}^e\big)}^{-1}. 
\end{equation}
}

\paragraph{Environment loss.} Denoted as $L_{env}$, guides the EBranch to learn external environment signals (aperiodic events). Our intuition is that the environment context of different samples from different time periods should be random and hence different (otherwise this becomes a periodic signal). 
To capture such varying environment contexts, the environment loss guides different samples to generate different environment representations. We randomly permute $\mathbf{g}^e$ along the batch dimension to obtain $\mathbf{g}^{e'}=\rm{\pi}(\mathbf{g}^e)$. We then obtain $B$ sample pairs  $(\mathbf{g}^e_i,\mathbf{g}^{e'}_i)$, $i\in [1, B]$. We use ${\rm softmax(\cdot)}$ to map $\mathbf{g}^e$ and $\mathbf{g}^{e'}$.
We aim to separate the representations of the sample pairs to guide \model\ to generate diverse environment representations for different times. Thus: 
{\small
\begin{equation}
    \label{eq:environment_loss}
    L_{env} = {\rm{KL}\big(\rm{\pi}(\mathbf{g}^e),\mathbf{g}^e\big)}^{-1}.
\end{equation}
}

\begin{figure}[t]
    \centering
    \includegraphics[width=0.95\linewidth]{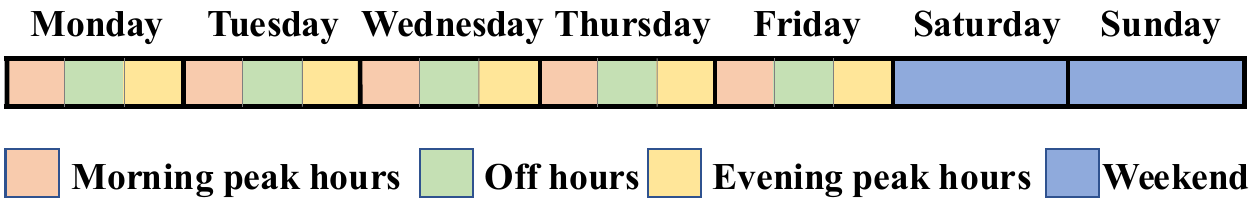}
    \caption{ Periodic patterns for the intrinsic branch.}   \label{fig:patterns}
\end{figure}

\paragraph{DBI loss.} Denoted as $L_{dbi}$ and inspired by the Davies–Bouldin index~(DBI)~\cite{dbi}, guides IBranch to learn representative intrinsic patterns.
Traffic observations exhibit different periodic patterns based on time (e.g., workdays vs. weekends, and peak hours vs. off hours).  We define 17 time-based patterns (Fig.~\ref{fig:patterns}): 15 for workdays (morning peak, off-hour, and evening peak for each weekday), one for Saturdays, and one for Sundays, due to the reduced variation on weekends~\cite{transgtr,dey2023traffic}. Public holidays are treated as Sundays.
The output $\mathbf{Z^i}$ of IBranch contains $B$ samples. We classify each sample based on its start time into one of the 17 patterns.  This gives a set of sample representations for each pattern. Let $P$ denote the set of 17 patterns, and $p \in P$ refer to one such set.
We define a matrix $\Psi \in \mathbb{R}^{|P| \times T \times N \times D}$ as the prototype for 17 patterns and optimise $\Psi$ during training. The intuition of  $\Psi$ is that each of 17 patterns may consist of $T$ distinct sub-patterns for each node, with each sub-pattern represented as a $D$-dimensional vector.
The DBI loss guides \model\ to learn more separated representations for the training samples with different periodic patterns, and closer representations for those with the same periodic patterns. 
We first compute two metrics $\mathcal{S}$ and $\mathcal{P}$ that evaluate the compactness of a pattern and the separation among patterns, respectively. 
{\small
\begin{equation}
    \label{eq:cp}
    \mathcal{S}_p(\mathbf{Z}^{i}, \Psi_p) = \frac{1}{|p|}\sum_{\mathbf{Z}^{i}_{j} \in p}||\Psi_p-\mathbf{Z}^{i}_{j}||_2.
\end{equation}
}
Here, $p$ is an element (also a set) of set $P$, $j$ is the $j$-th sample in $\mathbf{Z^i}$, and $\Psi_p$ denotes the slicing of $\Psi$ along the dimension of number of patterns that corresponds to $p$.  
Next, we compute $\mathcal{P}_{p,q}$ to evaluate the separation between sets $p, q \in P$, $\mathcal{P}_{p,q} = ||\Psi_p-\Psi_{q}||_2$. 
Another metric $\mathcal{R}_{p,q}$ balances the compactness of the two sets and the separation between them:
{\small
\begin{equation}
    \label{eq:r}
    \mathcal{R}_{p,q}(\mathbf{Z^i}, \Psi) = (\mathcal{S}_p + \mathcal{S}_q)\mathcal{P}_{p,q}^{-1}.
\end{equation}
}

Based on $\mathcal{R}_{p,q}(\mathbf{Z^i}, \Psi)$, we obtain a quality (in terms of  compactness and separation) score of set $p$, denoted by $\mathcal{D}_{p}$:
{\small
\begin{equation}
    \label{eq:D}
    \mathcal{D}_{p}(\mathbf{Z^i}, \Psi) = max_{p \neq q} \mathcal{R}_{p,q}.
\end{equation}
}

Finally, we can compute the DBI loss:
{\small
\begin{equation}
    \label{eq:dbi_loss}
    L_{dbi} = \frac{1}{|P|}\sum_{p \in P}\mathcal{D}_p(\mathbf{Z^i}, \Psi) = \frac{1}{|P|}\sum_{p\in P}\mathcal{D}_p(g_\omega^i(f_\theta^i(\mathbf{X^i})), \Psi).
\end{equation}
}

Based on the DBI loss, we can optimise prototype representations for each periodic pattern. We  enhance the representation $\mathbf{Z}=\rm{Concat}(\mathbf{Z^i},\mathbf{Z^e})$ by aggregating $\Psi$ as follows:
{\small
\begin{equation}
\label{eq:Z_Psi}
    \mathbf{Z}' = \mathbf{Z} + \mathbf{W} \Psi.
\end{equation}
}
Here, $\mathbf{W} \in \mathbb{R}^{B \times |P|}$ is a matrix where each row contains a one-hot vector indicating the pattern set to which each sample belongs, i.e., $w_{j,p} = 1$ if $\mathbf{Z^i}_j \in p$, otherwise  $w_{j,p} = 0$.
Based on Eq.~\ref{eq:Z_Psi}, we rewrite the prediction loss as follows:
{\small
\begin{equation}
    \label{eq:Z2X_hat}
    L_{pred} = \mathbb{E}(||g_\omega^{out}(\mathbf{Z'})-\mathbf{Y}||_p)
\end{equation}
}

\paragraph{Final loss.} Our final loss combines the three  loss terms above with the prediction loss (Eq.~\ref{eq:Z2X_hat}), weighted by hyper-parameters $\alpha$, $\beta$, and $\gamma$:
{\small
\begin{equation}\label{eq:loss_final}
L = L_{pred}+\alpha L_{flt} + \beta L_{env} + \gamma L_{dbi}.
\end{equation}
}
We include model time complexity in Appendix~\ref{subsec:a_dualcast_time_cpx}.

\subsection{Rooted Sub-tree Cross-time Attention}
\label{subsec:ct-attention}
The rooted sub-tree cross-time attention module consists of global and local attention mechanisms, which jointly capture high-order and dynamic spatial-temporal dependencies by learning correlations across time. This module is applied only within the spatial layers. For brevity, we omit the superscript `$sp$' in the notation. 
At each spatial layer, the input $\mathbf{H}_t^{l-1}$ (with $\mathbf{H}_t^{0} = \mathbf{X}_t$) is first projected into three subspaces to obtain the query $\mathbf{Q}^{l}$, key $\mathbf{K}^{l}$, and value $\mathbf{V}^{l}$. These representations are then used to compute the output $\mathbf{H}^l$. For simplicity, we omit the superscript `$l$' in the following notation.

Computing cross-time attention adds nodes from other time steps to the graph $G_{t}$ has $O(T^2N^2)$ time complexity when using scaled dot products as in prior models~\cite{gman,pdformer} (detailed in Appendix~\ref{subsec:a_TFM}). 
To reduce time complexity in cross-time attention, we first use a \emph{feature mapping function}~\cite{subtree-att}: 
{\small
\begin{equation}\label{eq:sp-sta-fm}
    \mathbf{h}_{t,n} = \frac{\phi(\mathbf{Q}_{t,n})\sum_{m=1}^N(\mathbf{M}_{n,m}\phi(\mathbf{K}_{t,m}))^{T}\mathbf{V}_{t,m}}{\phi(\mathbf{Q}_{t,n})\sum_{m=1}^N 
    \mathbf{M}_{n,m}\phi(\mathbf{K}_{t,m})^{T}},
\end{equation}
}where $\phi$ denotes a ReLU activation function; $n$ and $m$ are nodes in the graph; and $\mathbf{M}$ represents the connection (edge) between them.
The two summations terms $\sum_{m=1}^N \mathbf{M}_{n,m}\phi(\mathbf{K}_{t,m})^{T}$ and $\sum_{m=1}^N(\mathbf{M}_{n,m}\phi(\mathbf{K}_{t,m}))^{T}\mathbf{V}_{t,m}$ are shared by all nodes, which are computed once. Thus, we reduce the time complexity.

\begin{figure}[t!]
         \centering
  \includegraphics[width=\linewidth]{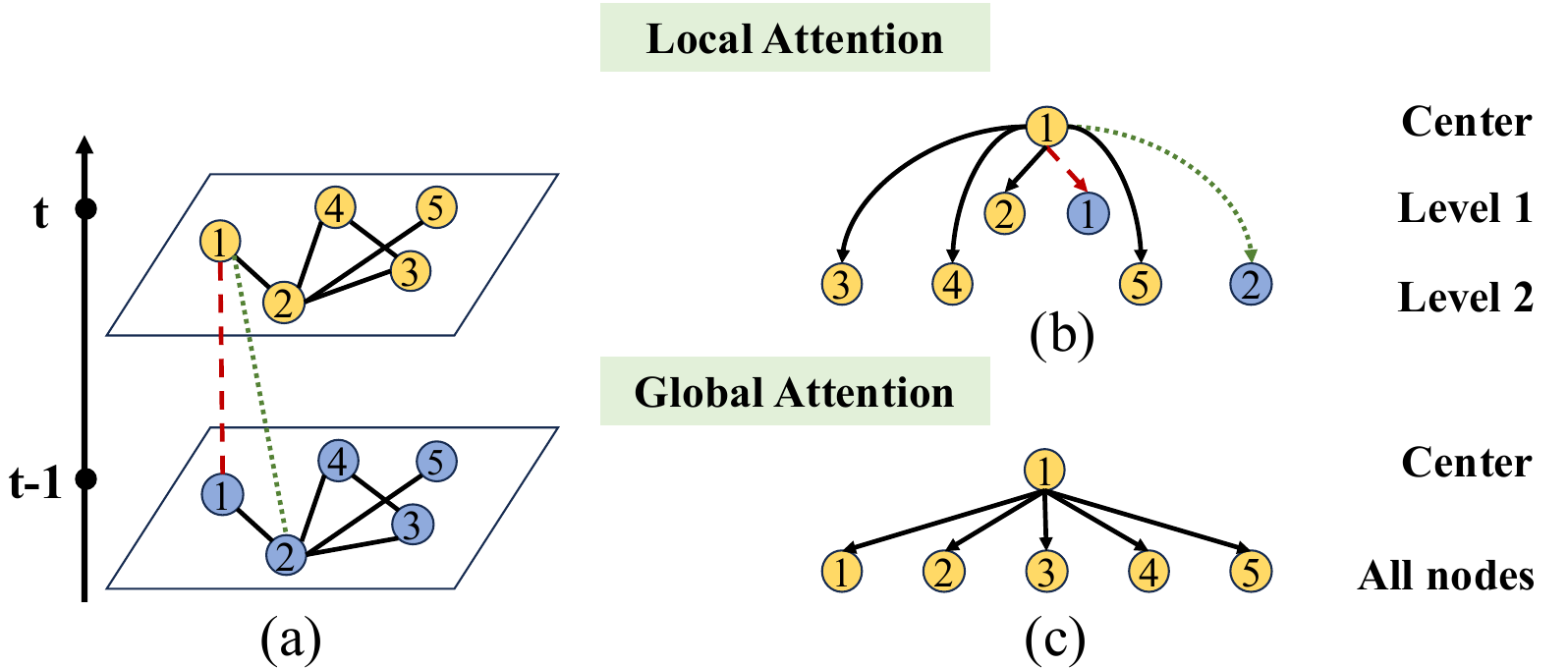}
         \caption{Rooted sub-tree cross-time attention. For simplicity, we set $t' - t'' = 1$. (a) Cross-time attention. Node \#1 at time step $t$ serves as the root of the subtree in (b). The red dashed line denotes an edge between observations of Node \#1 at different times. The green dotted line denotes an edge between Node \#1 and its one-hop neighbour at different times. The black lines denote edges between nodes and their one-hop neighbours observed at a time. (b) The subtree structure is used in local attention. (c) The global attention.}
         \label{fig:ct_att}
\end{figure}

\paragraph{Global attention.} We apply Eq.~\ref{eq:sp-sta-fm} to compute the self-attention among all nodes at time $t$, obtaining $N$ vectors $\mathbf{h}_{t,n}$ ($n\in [1,N]$), which form a global attention matrix $\mathbf{H}^{glo}_{t}$.
As Fig.~\ref{fig:ct_att}(c) shows, the global attention computes attention coefficients between all nodes (i.e., $\mathbf{M}$ in Eq.~\ref{eq:sp-sta-fm} is a matrix of 1's). We then update the representations of nodes by aggregating those from all other nodes, weighted by the attention coefficients. The time complexity of this process is $O(TN)$.

\paragraph{Local attention.} 
The local attention captures high-level correlations between nodes within a local area across different times. 
We achieve this goal by constructing an elaborate graph, where nodes from different times are connected. To learn high-level correlations, we reuse the feature mapping function-enhanced self-attention (Eq.~\ref{eq:sp-sta-fm}), instead of the stacked spatial-temporal GNN layers, for better efficiency and effectiveness. 

We use two types of edges to help learn cross-time correlations between nodes.
For a series of graph snapshots between times $t'$ and $t''$, we construct (1)~edges between the same node across different times (red dashed line in Fig.~\ref{fig:ct_att}(a)), and (2)~edges between a node and its one-hop neighbours from other times (green dotted line in Fig.~\ref{fig:ct_att}(a)).
This process yields a large cross-time graph with $N|t''-t'+1|$ nodes and edges from the original sensor graph at every time step, and the newly created edges.
We use the adjacency matrix of the cross-time graph as $\mathbf{M}$ in Eq.~\ref{eq:sp-sta-fm}, as visualised in Fig.~\ref{fig:ct-adj} (Appendix~\ref{subsec:a_rct}), where $\mathbf{A}^k$ is derived from $\mathbf{A}^1$ and $k$ indicates $k$-hop neighbours. 

A simple way to learn high-order relationships from graphs is to apply self-attention Eq.~\ref{eq:sp-sta-fm}, but it ignores the local hierarchical information. For example, the red dashed line and the green dotted line have different traffic propagation patterns and propagation time costs in Fig.~\ref{fig:ct_att}(a) as a red dashed line only concerns the same node across different times, while a green dotted line concerns nodes at different space and times, which should not be ignored. 

To fill this gap, we use a sub-tree structure that decouples the attention into multiple levels, as shown in Fig.~\ref{fig:ct_att}(b). This structure enables fine-grained control over the contributions from different hops within the graph. 
In the sub-tree construction process, red dashed lines represent the formation of 1-hop neighbours, while green dotted lines denote 2-hop neighbours. At each level $k$ ($k\in [1, \mathcal{K}]$, where $\mathcal{K}$ denotes the number of levels),
we compute the attention weights among the neighbours of a node $n$. These neighbours' representations are aggregated to obtain the localised representation $\mathbf{h}^{k,loc}_{t,n}$, as formalised in Eq.~\ref{eq:sp-sta-subtree}.
After computing $\mathbf{h}^{k,loc}_{t,n}$ for all nodes, the resulting vectors are assembled into the matrix $\mathbf{H}^{k,loc}_{t}$. 
Subsequently, we aggregate representations from all levels to form the final local attention $\mathbf{H}^{loc}_{t}=\sum_{k=0}^{\mathcal{K}}w_{k}\mathbf{H}^{k,loc}_{t}$. Here, $w_{k}$ is a learnable parameter to control the contribution of each hop. This computation process can be seen as a $k$-hop message-passing process. Based on the $k$-hop message-passing process, the mask $\mathbf{M}^{k}$ in each step equals to $\mathbf{A}^1$, denoted as $\mathbf{A}$. Fig.~\ref{fig:ct-adj}(b) shows this matrix, where $I_N$ denotes an identity matrix of size $N$. Since the message-passing process runs for each edge and among all nodes, we obtain $\mathbf{H}^{loc}_{t}$ with time complexity $O(|E'|)$. Here, $E'$ represents the number of edges after we build the edges across graphs from different time steps (i.e., number of $1$'s in $\mathbf{A}$). 
{\small
\begin{flalign}
\mathbf{h}^{0,loc}_{t,n} = \mathbf{V}_{t,n},
\nonumber
\end{flalign}
\vspace{-0.5cm}
\begin{align}
\label{eq:sp-sta-subtree}
    \mathbf{h}^{k,loc}_{t,n} = \frac{\phi(\mathbf{Q}_{t,n})\sum_{m=1}^N(\mathbf{M}^{k}_{n,m}\phi(\mathbf{K}_{t,m}))^{T}\mathbf{V}_{t,m}}{\phi(\mathbf{Q}_{t,n})\sum_{m=1}^N 
    \mathbf{M}_{n,m}^{k}\phi(\mathbf{K}_{t,m})^{T}}.
\end{align}
}

After obtaining $\mathbf{H}^{loc}_{t}$ and $\mathbf{H}^{glo}_{t}$, we fuse them as follows:
{\small
\begin{equation}
    \mathbf{H}_{t} = \mathbf{H}^{loc}_{t} + w_{glo}\mathbf{H}^{glo}_{t},
\end{equation}
}where $w_{glo}$ is a learnable parameter. Then, we concatenate $\mathbf{H}_{t}$ from each time step $t$ to obtain the output of the spatial self-attention layer $\mathbf{H}= ||_{t=0}^{T} \mathbf{H}_{t}$.

We also use a temporal self-attention module to capture the temporal features from the full input time window. We merge $\textbf{H}^{sp}$ with $\textbf{H}^{te}$ to obtain the final output of each layer.

\paragraph{Discussion.}
The high-order and dynamic spatial-temporal relationships play an important role in traffic forecasting. Previous graph-based methods~\cite{stfgnn,stsgcn} stack GNNs to capture such correlations, with sub-optimal effectiveness, while the vanilla self-attention models suffer in their quadratic time complexity. 
Our work addresses these issues and presents a versatile self-attention-based method to exploit the high-order and dynamic spatial-temporal relationships effectively and efficiently. 

\begin{table*}[!ht]
\small
\centering
 \resizebox{\textwidth}{!}{
 \setlength\columnsep{0pt}
\begin{tabular}
{l|cc|cc|cc}
  \hlineB{3}
  \multirow{2}{*}{\textbf{Model}} & \multicolumn{2}{c|}{\textbf{PEMS03}}& \multicolumn{2}{c|}{\textbf{PEMS08}}& \multicolumn{2}{c}{\textbf{Melbourne}}\\
  \cline{2-7}
  & RMSE$\downarrow$  & MAE$\downarrow$  & RMSE$\downarrow$  & MAE$\downarrow$  & RMSE$\downarrow$  & MAE$\downarrow$\\
  \hline
  \hline
  GWNet	& 26.420$\pm$0.839	& 15.404$\pm$0.052	& 25.796$\pm$0.318	& 16.314$\pm$0.230 &	25.778$\pm$0.136 & 13.410$\pm$0.065\\
  MTGNN	& \underline{25.413}$\pm$0.201	& 14.707$\pm$0.070 & 23.794$\pm$0.073 & 14.898$\pm$0.066 & 25.364$\pm$0.291 & 13.310$\pm$0.137\\
  STPGNN & 25.889$\pm$0.718 & 14.868$\pm$0.141& 23.374$\pm$0.088 & 14.202$\pm$0.085 & 25.170$\pm$0.083 & 13.075$\pm$0.071\\
  MegaCRN & 25.645$\pm$0.119 & 14.733$\pm$0.031 & 24.052$\pm$0.333 & 15.118$\pm$0.034 & \underline{24.482}$\pm$0.143 & \underline{12.647}$\pm$0.033\\
 EASTNet & 26.920$\pm$1.732 & 16.356$\pm$1.511 & 27.166$\pm$2.088 & 17.574$\pm$1.910 & 28.494$\pm$1.143 & 15.310$\pm$0.746\\
 STNorm & 27.328$\pm$0.437 & 16.382$\pm$0.191 & 26.026$\pm$0.119 & 17.090$\pm$0.113& 25.744$\pm$0.277 & 13.724$\pm$0.134\\
 STWave	& 26.346$\pm$0.174 & 14.975$\pm$0.086 & 23.270$\pm$0.108& \underline{13.480}$\pm$0.089 & 26.918$\pm$0.455 & 14.060$\pm$0.341\\
  STTN & 27.166$\pm$0.408	& 15.490$\pm$0.115 & 24.984$\pm$0.248 & 15.924$\pm$0.200 & 26.402$\pm$0.247 & 13.790$\pm$0.117\\
  GMAN & 26.624$\pm$0.503 & 15.520$\pm$0.111 & 23.750$\pm$0.194 & 14.080$\pm$0.036 & 24.496$\pm$0.351 & 12.652$\pm$0.136\\
  PDFormer & 25.950$\pm$0.421 & \underline{14.690}$\pm$0.080 & \underline{23.250}$\pm$0.099 & 13.654$\pm$0.337 & 27.072$\pm$0.301 & 14.230$\pm$0.069\\
  \hline\hline
  \model-S (ours) & 26.282$\pm$0.128 ({\color{blue}+3.3\%}) & 15.370 $\pm$0.086 ({\color{blue}+0.8\%}) & 24.526$\pm$0.116 ({\color{blue}+1.8\%}) & 15.550$\pm$0.119 ({\color{blue}+2.3\%}) & 25.474 $\pm$0.184 ({\color{blue}+3.5\%})& 13.316$\pm$0.061 ({\color{blue}+3.4\%})\\
  \model-G (ours) & 25.582$\pm$0.414({\color{blue}+3.9\%}) & 15.094$\pm$0.099 ({\color{blue}+2.7\%}) & 23.564$\pm$0.133 ({\color{blue}+0.8\%}) & 13.938$\pm$0.114 ({\color{blue}+1.0\%}) & \textbf{23.978}$\pm$0.105 ({\color{blue}+2.1\%}) & \textbf{12.420}$\pm$0.057 ({\color{blue}+1.8\%})\\
  \model-P (ours) & \textbf{24.898}$\pm$0.663 ({\color{blue}+4.1\%})& \textbf{14.666}$\pm$0.087 ({\color{blue}+0.2\%}) & \textbf{22.998}$\pm$0.161 ({\color{blue}+1.1\%}) & \textbf{13.332}$\pm$0.059  ({\color{blue}+2.4\%}) & 26.040$\pm$0.150 ({\color{blue}+3.8\%}) & 13.542$\pm$0.003 ({\color{blue}+4.8\%})\\
  \hline
  \rowcolor{gray!20}
  Error reduction & 2.0\% & 0.2\% & 1.1\%& 1.1\%& 2.1\% & 1.8\%\\
  \hlineB{3}
\end{tabular}
}
\caption{Overall model performance. ``$\downarrow$''  indicates lower values are better. The best baseline results are \underline{underlined}, and the best \model\ results are in boldface. 
``Error reduction''  denotes the percentage decrease in errors of the best \model-based model compared to the best baseline.
{The numbers in {\color{blue}blue} show error reduction achieved by a \model-based model over vanilla models, e.g., \model-\underline{P} vs. \underline{P}DFormer.}}
\label{tab:overallresults}
\end{table*}

\section{Experiments}
\subsection{Experimental Setup}
\textbf{Datasets.}
We use two freeway traffic datasets and an urban traffic dataset: 
\textbf{PEMS03} and \textbf{PEMS08}~\cite{pems} contain traffic flow data collected by 358 and 170 sensors on freeways in California;  \textbf{Melbourne}~\cite{STSM} contains traffic flow data collected by 182 sensors in the City of Melbourne, Australia.
The traffic records in PEMS03 and PEMS08 are given at 5-minute intervals (288 intervals per day), while those in Melbourne are given at 15-minute intervals (96 intervals per day). Melbourne has a higher standard deviation and is more challenging. See Table~\ref{tab:dataset_info} (Appendix~\ref{subsec:a_exp_setup}) for the dataset statistics.

Following~\citeauthor{dcrnn}, we use records from the past hour to forecast for the next hour, i.e., $T=T'=1\ hour$ in Eq.~\ref{eq:pred_loss} over all datasets. We split each dataset into training, validation, and testing sets by 7:1:2 along the time axis. 

\paragraph{Competitors.} 
\model\ works with spatial-temporal models that follow the described self-attention-based structure. We implement \model\ with three such models: \textbf{GMAN}~\cite{gman}, \textbf{STTN}~\cite{sttn}, and \textbf{PDFormer}~\cite{pdformer}, denoted as \textbf{\model-G}, \textbf{\model-S}, and \textbf{\model-P}, respectively. 
We compare with the vanilla GMAN, STTN, and PDFormer models, plus GNN-based models \textbf{GWNet}~\cite{graphwavenet}, \textbf{MTGNN}~\cite{MTGNN}, and \textbf{STPGNN}~\cite{STPGNN}. We further compare \model\ with \textbf{MegaCRN}~\cite{megaCRN} and \textbf{EAST-Net}~\cite{eastnet}, which focus on modelling non-stationarity in spatial-temporal series. We also compare with \textbf{STWave}~\cite{stwave} and \textbf{STNorm}~\cite{stnorm}, which consider disentanglement in forecasting.

\paragraph{Implementation details.}
We use the public released code of the competitors, except for STTN which is implemented from Libcity~\cite{libcity}. We implement \model\ with the self-attention-based models following their source code, using PyTorch. 
We use the default settings from the source code for both the baseline models and their variants powered by \model. 
We train the models using Adam with a learning rate starting at 0.001, each in 100 epochs. 
For the models using \model, we use grid search on the validation sets to tune the hyper-parameters $\alpha$, $\beta$, and $\gamma$. 
Table~\ref{tab:parameters} (Appendix~\ref{subsec:a_exp_setup}) lists these hyper-parameter values. 
All experiments are run on an NVIDIA Tesla A100 GPU with 80 GB RAM. Following~\citeauthor{cast}, we use the average of root mean squared errors~(\textbf{RMSE}) and mean absolute errors~(\textbf{MAE}) for evaluation. Results are averaged over five runs.

\subsection{Overall Results}
\label{subsec:exp_results}

\begin{figure*}[!t]
    \centering
    \begin{subfigure}[b]{0.33\linewidth}
        \includegraphics[height=3.6cm]{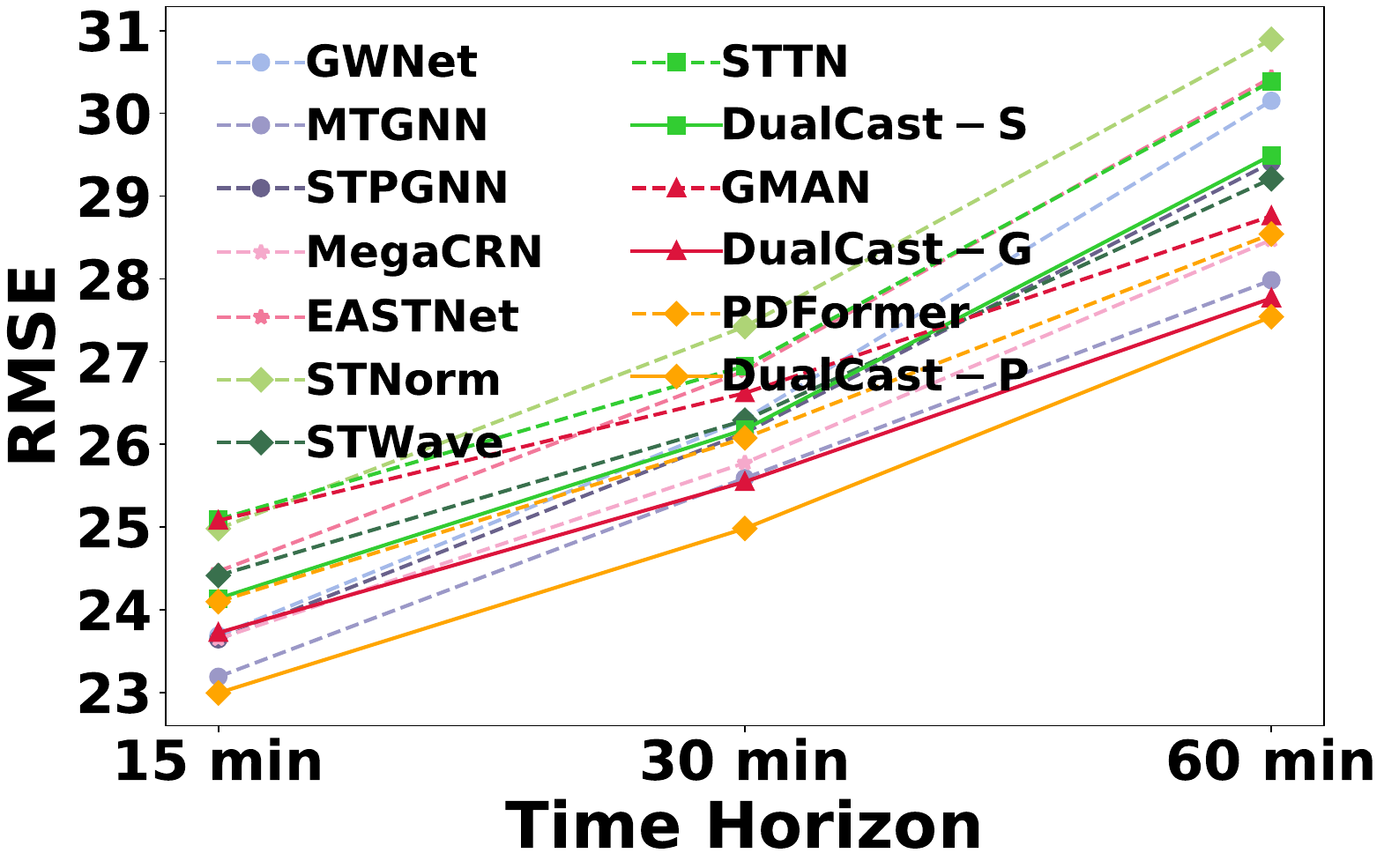}
    \caption{\footnotesize Forecasting error vs. time horizon}
    \label{subfig:rmse_horizon}
    \end{subfigure}
    \begin{subfigure}[b]{0.33\linewidth}
        \includegraphics[height=3.6cm]{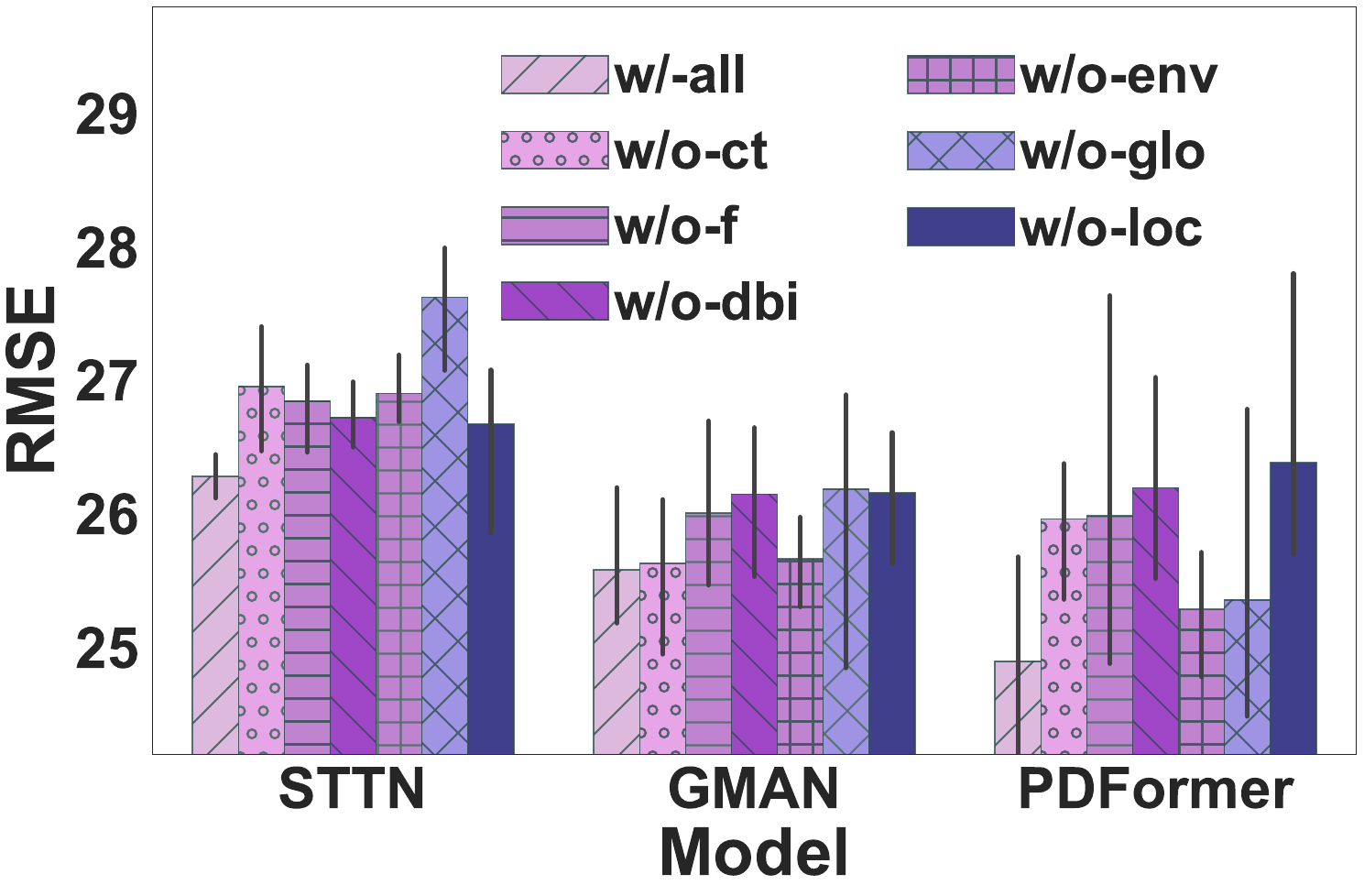}
    \caption{\footnotesize Ablation Study}
    \label{subfig:ablation_study}
    \end{subfigure}
    \begin{subfigure}[b]{0.33\linewidth}
        \includegraphics[height=3.6cm]{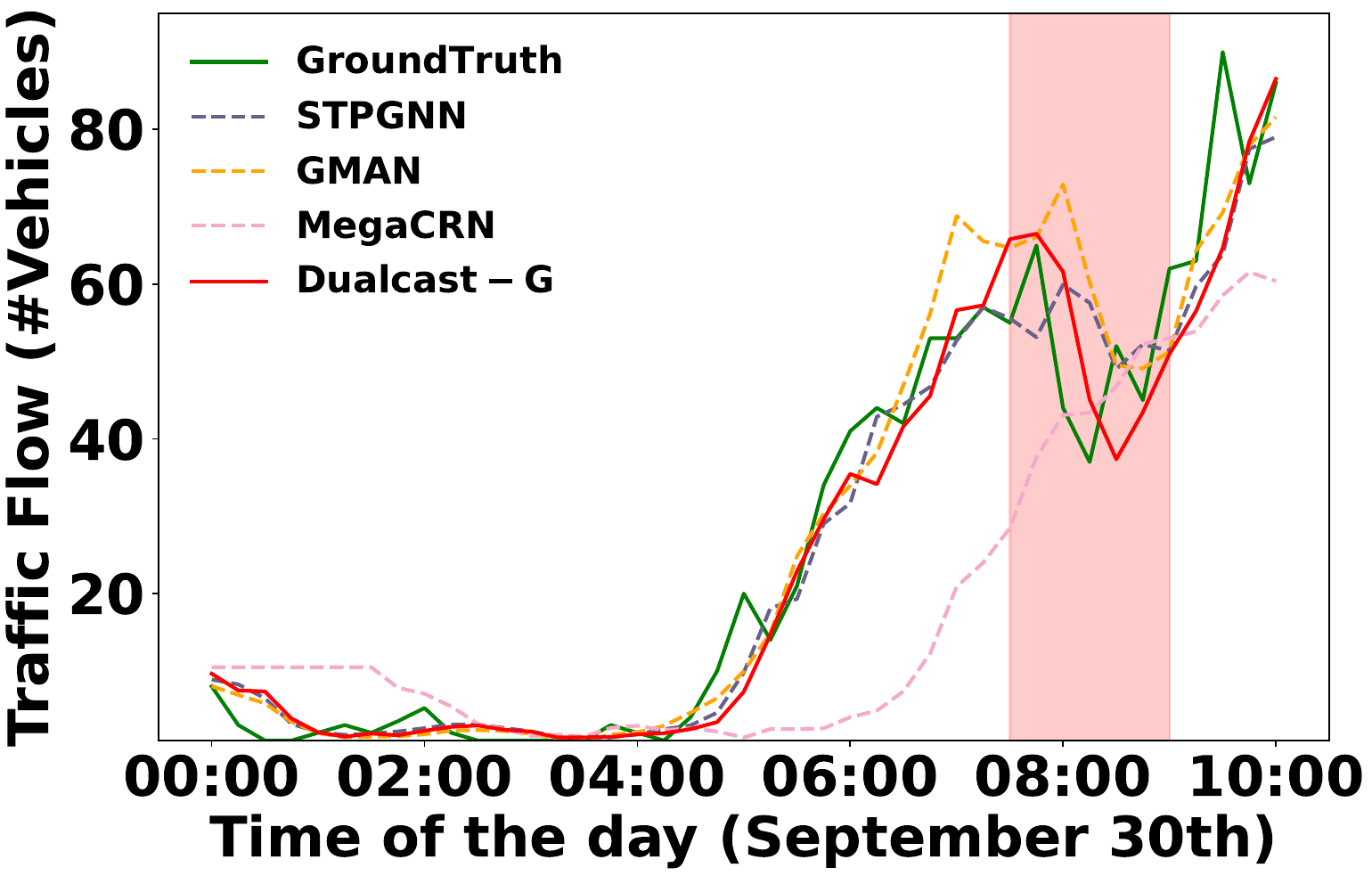}
    \caption{\footnotesize Car accident}
    \end{subfigure}
    \caption{\small Results on PEMS03 (PEMS08 and Melbourne results in Appendix~\ref{subsec:a_over_results} to~\ref{subsec:a_param_study}).~(a) shows forecasting error vs. time horizon, with dashed lines for baseline models and solid lines for \model-based models. The {\color{green}green}, {\color{red}red}, and {\color{orange}orange} lines represent \model-S, \model-G and \model-P, respectively, with their counterparts. (b) lists Ablation study results on PEMS03. (c) presents a case study for a car crash at sensor \#138 in Melbourne with the occurrence time marked in red.}
    \label{fig:model_performance}
\end{figure*}

\paragraph{Model performance across all times.} Table~\ref{tab:overallresults} reports forecasting errors averaged over one hour.  Powered by \model, \model-G, \model-P, and \model-S consistently outperform their vanilla counterparts. \model-P has the best performance on freeway traffic datasets PEMS03 and PEMS08, while \model-G performs the best on the urban traffic dataset Melbourne. Compared to PEMS03 and PEMS08, Melbourne represents an urban environment with greater variability in the traffic flow series (Table~\ref{tab:dataset_info}). GMAN's simple, robust design explains its strong performance in Melbourne, while PDFormer's reliance on time series clustering excels on PEMS03 and PEMS08 but struggles with Melbourne's variability, hindering cluster formation and accuracy. 
Fig.~\ref{subfig:rmse_horizon} further shows the RMSE for forecasting 15, 30, and 60 minutes forecasts, comparing \model-G, \model-P, and \model-S, their vanilla counterparts and top baselines MegaCRN and STPGNN.
We see that the \model\ models outperform the baseline models consistently at different forecasting horizons, confirming their effectiveness. We also conducted t-tests and Wilcoxon tests over our model and the best baseline models across all datasets. We find that all results are statistically significant (with $p \ll 10^{-8}$).

\paragraph{Model performance during hours prone to traffic accidents.}
To verify \model's ability to learn complex environment contexts, we examine forecasting results during \emph{complex times} (4:00 pm to 8:00 pm on workdays), which has reported a higher chance of traffic accidents~\cite{carsh_freq1,carsh_freq2}.
Table~\ref{tab:framework_with_dualcast} shows the results, where ``all'' means the RMSE at the 1-hour horizon for all days and ``cpx'' means that at complex times. 

All models, including ours, have larger errors at complex times in most cases (except for STWave, GMAN, \model-G, and \model-P on Melbourne), confirming it as a challenging period. Importantly, the errors of the \model-based models increase less at complex times compared to their vanilla counterparts.  For example, the error gaps between \model-G and GMAN, and \model-P and PDFormer on PEMS08 double from 1.6\% to 3.0\% and 0.7\% to 1.4\%, respectively. Meanwhile, \model-P reduces the errors by up to 9.6\% in Melbourne. These results confirm that disentangling intrinsic and environmental contexts improves forecast accuracy. Exceptions on Melbourne may arise due to its high data complexity, variance and skewness, making even a less complex time challenging.

\begin{table*}[!t]
\tiny
\centering
\resizebox{\textwidth}{!}{
\begin{tabular}{l|cc|cc|cc}
\hlineB{3}
\multicolumn{1}{c|}{}                                        & \multicolumn{2}{c|}{\textbf{PEMS03}}                                            & \multicolumn{2}{c|}{\textbf{PEMS08}} & \multicolumn{2}{c}{\textbf{Melbourne}} \\ \cline{2-7} 
\multicolumn{1}{c|}{\multirow{-2}{*}{\textbf{Model}}}        & \textbf{all}                           & \textbf{cpx}                           & \textbf{all}      & \textbf{cpx}     & \textbf{all}       & \textbf{cpx}      \\ 
\hline
\hline
GWNet                                                       & 30.152                                 & 37.276                                 & 29.162            & 31.986           & 27.766             & 29.186            \\
MTGNN                                                       & \underline {27.984}                           & \underline {34.842}                           & 26.253            & 28.616           & 27.264             & 27.994            \\
STPGNN                                                      & 29.405                                 & 35.346                                 & 26.095            & 27.320           & 27.018             & 27.510            \\
MegaCRN                                                     & 28.472                                 & 35.692                                 & 27.110            & 27.442           & 26.360             & 26.654            \\
EASTNet                                                    & 30.434                                 & 37.786                                 & 31.066            & 32.674           & 31.240             & 33.220            \\
ST-Norm                                                     & 30.895                                 & 38.956                                 & 29.366            & 31.948           & 27.752             & 29.304            \\
STWave                                                      & 29.210                                 & 37.312                                 & 25.600            & 27.130           & 29.562             & 29.324            \\
STTN                                                        & 30.386                                 & 40.990                                 & 28.182            & 31.400           & 29.108             & 30.554            \\
GMAN                                                        & 28.760                                 & 36.078                                 & 25.632            & \underline {26.886}     & \underline {26.066}       & \underline {25.045}      \\
PDFormer                                                    & 28.542                                 & 35.416                                 & \underline {25.468}      & 27.022           & 30.048             & 30.060            \\ \hline
DualCast-S (ours)                                                  & 29.494 ({\color{blue}+2.9\%}) & 39.138 ({\color{blue} \textbf{+4.5}\%}) & 27.522 ({\color{blue} +2.3\%})   & 30.454 ({\color{blue} \textbf{+3.0}\%})   & 27.256 ({\color{blue} \textbf{+6.4}\%})    & 28.946 ({\color{blue} +5.3\%})   \\
DualCast-G (ours)                                                  & 27.766 ({\color{blue} +3.5\%})                        & 34.450 ({\color{blue} \textbf{+4.5}\%})                         & 25.464 ({\color{blue} +0.7\%})   & 26.506 ({\color{blue} \textbf{+1.4}\%})  & \textbf{25.468} ({\color{blue} +2.3\%})    & \textbf{24.398 }({\color{blue} \textbf{+2.6}\%})   \\
DualCast-P (ours)                                                  & \textbf{27.542} ({\color{blue} +3.5\%})                        & \textbf{33.950} ({\color{blue} \textbf{+4.7}\%})                         & \textbf{25.048} ({\color{blue} +1.6\%})   & \textbf{26.224} ({\color{blue} \textbf{+3.0}\%})  & 28.134 ({\color{blue} +6.4\%})    & 27.168 ({\color{blue} \textbf{+9.6}\%})   \\ 
\hline
\rowcolor{gray!20}
Error reduction & 1.6\%                                 & {2.6}\%                                 & 1.7\%            & {2.5}\%           & 2.3\%             & {2.6}\%    \\
\hlineB{3}
\end{tabular}
}
\caption{Model performance (RMSE at the 1-hour horizon) for ``all'' times (any time of day) and ``cpx'' times (4:00 pm to 8:00 pm with complex contexts).
 Best baseline results are \underline{underlined}, and \model's best results are in boldface. {\color{blue}blue} numbers show error reduction by \model-based models compared to their vanilla counterparts.}
 \label{tab:framework_with_dualcast}
\end{table*}

\subsection{Ablation Study}
We implement six model variants:~\textbf{w/o-ct} disables the CT attention from \model;  \textbf{w/o-f}, \textbf{w/o-dbi}, and \textbf{w/o-env} remove the filter loss, the DBI loss, and the environment loss, respectively; \textbf{w/o-glo} and \textbf{w/o-loc} remove the global attention and local attention (including the CT attention), respectively. 

As Fig.~\ref{subfig:ablation_study} shows, all \model\ modules enhance model performance. The DBI loss is more important on PEMS03, as PEMS03's freeway data exhibits clearer patterns across times and days, which \model\ can effectively learn with DBI loss guidance. More results are in Appendix~\ref{subsec:a_ablation_study}.

\subsection{Parameter and Case study}
\paragraph{Parameter study.}
We study the impact of \model\ hyper-parameters in our loss function (Eq.~\ref{eq:loss_final}), namely $\alpha$, $\beta$, and $\gamma$. Results confirm that \model\ is robust without the need for heavy tuning. More details are in Appendix~\ref{subsec:a_param_study}.

\begin{figure}[t]
         \centering
  \includegraphics[width=0.5\textwidth]{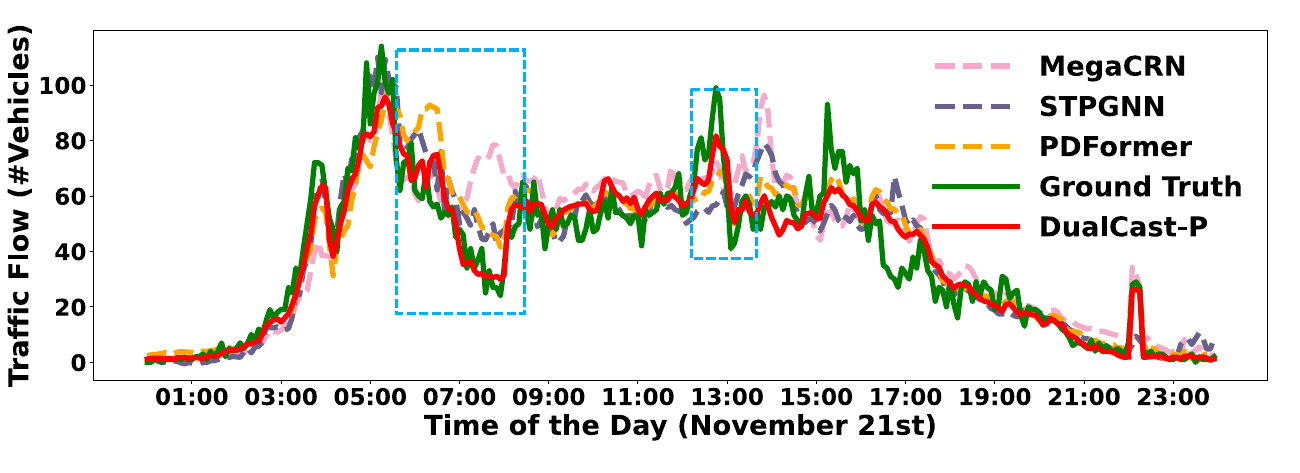}
         \caption{A case study of responding to sudden changes (highlighted in rectangles) in traffic at sensor \#97 on PEMS03 on Nov. 21.}
         \label{fig:sudden_change}
\end{figure}

\paragraph{Case study.}
We conduct two case studies below.

\emph{Responding to traffic accidents.} 
Cross-referencing Melbourne car crash reports~\cite{vic_car_crash} with the dataset revealed one accident during the data period. Fig.~\ref{fig:model_performance}(c) compares 15-minute-ahead forecasts from \model-G (best on this dataset), its vanilla counterpart GMAN, and the top-2 baselines (STPGNN, MegaCRN) at the sensor nearest the accident.
Around 8:00, \model-G quickly captures the traffic change caused by the crash, producing forecasts closest to the ground truth.

{\emph{Responding to sudden changes in traffic.}} 
On the PEMS datasets, ground-truth traffic events are unavailable. Instead, we found two representative sensors ($\#72$, see Appendix~\ref{subsec:a_param_study} and $\#97$) on PEMS03 with sudden changes on November 21st, 2018.
Fig.~\ref{fig:sudden_change} shows the ground-truth traffic flow and 1-hour-ahead forecasts by MegaCRN, STPGNN, PDFormer, and \model-P at Sensor $\#97$. \model-P closely aligns with the ground truth, especially during sudden changes, again highlighting the strength of \model. 
More results are in Appendix~\ref{subsec:a_over_results}~to~\ref{subsec:a_param_study}, including PEMS08 results, an analysis of model scalability, effectiveness and potential for GNN-based models, and visualisations for disentangling results.

\section{Conclusion}
We proposed a framework named \model\ that enhances the robustness of self-attention-based traffic forecasting models, including the SOTA, in handling scenarios with complex environment contexts. \model\ takes a dual-branch structure to disentangle the impact of complex environment contexts, as guided by three loss functions. We further proposed a cross-time attention module to capture high-order spatial-temporal relationships. We performed experiments on real-world freeway and urban traffic datasets, where models powered by \model\ outperform their original versions by up to 9.6\%. The best \model-based model outperforms the SOTA model by up to 2.6\%, in terms of forecasting errors. 

\section*{Ethical Statement}
All datasets used in this study are publicly available and do not contain any personally identifiable information. There are no ethical concerns associated with this work.

\section*{Acknowledgments}
This work is in part supported by the Australian Research Council (ARC) via Discovery Projects DP230101534 and DP240101006. Jianzhong Qi is supported by ARC Future Fellowship FT240100170. Feng Liu is supported by the ARC with grant numbers DE240101089, LP240100101, DP230101540 and the NSF\&CSIRO Responsible AI program with grant number 2303037.

\bibliographystyle{named}
\bibliography{ijcai25}

\newpage
\appendix
\section{Related Work}
\label{sec:a_related_work}
We summarise the relevant literature on traffic forecasting models and how existing solutions (1) model correlations between different locations and times, (2) address aperiodic events, and (3) disentangle traffic series. 

\paragraph{Traffic forecasting models.}
Traffic forecasting is a spatial-temporal forecasting problem. Initially, traffic forecasting studies focused on temporal pattern modelling~\cite{arima_traffic}, employing statistics time-series forecasting models such as ARIMA~\cite{arima}. To better capture temporal patterns, more advanced models, such as the recurrent neural network (RNN) and its variants, are used in follow-up studies~\cite{tian2015predicting,lstm_traffic}. 

Dynamic spatial correlations are considered in later studies. 
Since road networks can be represented as graphs, graph neural networks (GNNs)-based models~\cite{dcrnn,stgnn,stfgnn,astgcn,gcnode,stsgcn,graphwavenet,gamcn,sttn,MTGNN,STPGNN} have recently become dominant in traffic forecasting. These studies typically adopt GNNs to capture spatial features, along with RNNs~\cite{dcrnn} or 1-D temporal convolution models (TCNs)~\cite{graphwavenet,gamcn,stfgnn,gcnode} to capture temporal features. Compared with RNNs, TCNs are more efficient and can alleviate gradient explosion (or vanishing) issues in long-term forecasting.

Beyond spatial correlations, the semantic correlations between sensors also attract much attention. Different methods have been proposed to build the adjacency matrix, such as using road network distances~\cite{stgcn,jin2022selective}, node embedding similarity~\cite{graphwavenet,gamcn,megaCRN}, time-series similarity~\cite{STSM}, and region similarity~\cite{increase,jin2022selective,hafusion,FlexiReg}. However, these adjacency matrices can only model static relationships, while real traffic patterns (e.g., traffic flows between residential and industrial areas) may vary across time.
To address this gap, the attention mechanism is used in recent models to capture further dynamic (i.e., time-varying) relationships~\cite{gman,pdformer,sttn,astgcn,staeformer,stabc} and has achieved strong outcomes. Learning different spatial correlations for different periods, such as one adjacency pattern for workday peak hours and another for weekends, is another  solution~\cite{ma2024learning}. 

\paragraph{Modelling correlations between different locations at different times.} 
Though the models above are effective for capturing periodic patterns, they struggle with modelling aperiodic events and the ripple effect of these events, because they model the spatial and temporal correlations separately. To adapt to aperiodic events, a few GCN-based models~\cite{stfgnn,cast,hyTemporalGraph} connect graph snapshots captured at different time steps to learn correlations between different sensor locations at different times. Such models treat impact of nodes from different hops equally, which does not reflect real-world observations where traffic propagation between nodes at different distances takes different times. 
Therefore, they still struggle with capturing the propagation patterns of aperiodic events. Our proposed cross-time attention learns different weights to model such varying impacts. 


\paragraph{Aperiodic event-aware traffic forecasting.} Aperiodic events, such as traffic incidents and social events, are common in real-world (spatial-temporal) traffic series. 
Accurate forecasting with such events is important for downstream applications, e.g., transportation optimisation and route planning. However, most existing models have poor generalisation of aperiodic events.
 East-Net~\cite{eastnet} and MegaCRN~\cite{megaCRN} adopt memory augment to alleviate this issue, where the memory augment (i.e., memory network) enhances model representations by learning prototype representations of spatial-temporal patterns, including aperiodic patterns.  
Though such memory networks can enhance model generalisation, it is difficult to predefine the number of memories to model different periodic and aperiodic patterns. Our proposed \model\ utilises environment loss to encourage the model to learn diverse aperiodic patterns without predefined numbers of aperiodic patterns, which is more flexible.

\paragraph{Disentangling traffic series.} 
Real-world traffic series are complex and entangle multiple kinds of signals. Disentangling traffic series is an intuitive yet non-trivial task for more accurate traffic forecasting. Previous study denoises traffic data~\cite{chen2021traffic}, while the noise could contain useful information, such as a sudden change caused by a traffic incident --  ignoring such ``noise'' limits model generalisation. MUSE-Net~\cite{museNet} focuses on the disentanglement of multiple periodic signals rather than aperiodic signals. TimeDRL~\cite{TimeDRL} also uses a dual-branch structure, where one learns representations at the time-stamp level (i.e., each time step in the input window has an embedding), and the other at the instance level (i.e., each input window has an embedding). A couple of other studies extract different traffic modes from traffic data using wavelet transform and decoupling~\cite{modwavemlp,stwave}. This approach struggles when encountering long-term aperiodic events, such as serious car accidents or major social events, which may confuse the predefined wavelets. 

Disentangling and fusing high-frequency and low-frequency signals have been used to enhance model performance~\cite{stnorm,fouriergnn}. 
More recent studies disentangle traffic series into invariant signals~\cite{cast,zhou2023maintaining} and environment signals, formed by a predefined set of city zones (e.g., commercial vs. residential) or environmental factors. 
These models are not based on self-attention, while we do. More importantly, our environment loss functions together with a dual-branch model structure enable capturing environment contexts with a higher flexibility, without the need for predefined aperiodic patterns.

\section{Additional Details on \model\ Design}
\label{sec:a_method}

\subsection{Self-Attention-Based Traffic Forecasting}
\label{subsec:a_TFM}
We describe typical structures followed by self-attention-based traffic forecasting (\emph{STF} hereafter) models~\cite{pdformer,gman,sttn,STPGNN}. Our \model\ framework and attention-based optimisations are designed to be compatible with such structures.  

STF models use stacked \emph{spatial-temporal layers} (STLayers) as an \emph{encoder} $f_\theta$ to capture spatial-temporal correlations, with the learned representations passed to a \emph{decoder} $g_\omega$ for forecasting.

\begin{figure}[t]
         \centering
  \includegraphics[width=0.85\linewidth]{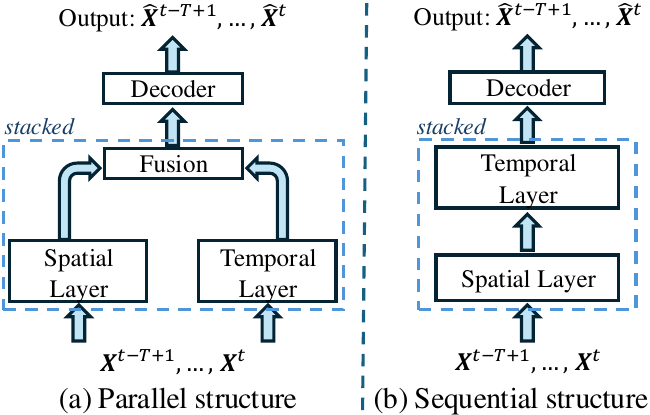}
         \caption{Spatial-temporal forecasting model structures}
         \label{fig:stf-model}
\end{figure}

\paragraph{Spatial-temporal layer.}
Two structures are commonly used for the STLayers, i.e., parallel and sequential, as illustrated by Fig.~\ref{fig:stf-model}.
Parallel STLayers learn spatial and temporal features separately and then fuse the hidden representations of the two types of features to form the spatial-temporal features.
In contrast, sequential STLayers process the two types of features one after another.
Multiple STLayers are often stacked to learn richer spatial-temporal correlations.

\paragraph{Spatial layer.} 
Spatial layers (SLayers) focus on the spatial correlations between sensors.
The $l$-th SLayer first projects the input $\mathbf{H}^{l-1}_{t}$ (and $\mathbf{H}^{0}_{t} = \mathbf{X}_{t}$) at time $t$ into three sub-spaces to get query $\mathbf{Q}_{t}^{sp,l}$, key $\mathbf{K}_{t}^{sp,l}$ and value $\mathbf{V}_{t}^{sp,l}$ in dimensionality $d$.
Then, we compute $\mathbf{h}^{sp,l}_{t,n} \in \mathbb{R}^{d}$, i.e., the hidden representation for the node $n$ at time $t$: 
{\small
\begin{equation}
\label{eq:sp-sta-mask}
\mathbf{h}^{sp,l}_{t,n} = \sum_{m=1}^{N}\mathrm{softmax} (\mathrm{sim}(\mathbf{Q}_{t,n}^{sp,l},{\mathbf{K}_{t,m}^{sp,l}}^{T})\odot \mathbf{M}^{sp}_{n,m}) \mathbf{V
}_{t,m}^{sp,l}.
\end{equation}
}
\hspace*{-2mm}Here, $\mathrm{sim}(\mathbf{X}, \mathbf{Y}) = \frac{\mathbf{X}\mathbf{Y}}{\sqrt{d}}$ is a function to compute the similarity between two matrices. 
$\mathbf{M}^{sp}_{n,m}$ represents the mask between node $n$ and node $m$; ~\cite{gman} and ~\cite{pdformer} used different kinds of $\mathbf{M}^{sp}$ as mention in Section~\ref{sec:related_work}.
The $l$-th SLayer applies the self-attention across all nodes, yielding $N$ vectors $\mathbf{h}^{sp,l}_{t,n}$ $(n \in [1,N])$, which together form matrix $\mathbf{H}_{t}^{sp,l} \in \mathbb{R}^{N\times D}$.
Then, the $l$-th SLayer applies the self-attention at each time step and concatenates the output from each time step to produce the output of the layer $\mathbf{H}^{sp,l} = \mathrm{concat}(\mathbf{H}_{t-T+1}^{sp,l}, \ldots, \mathbf{H}_{t}^{sp,l})$.

\paragraph{Temporal layer.} 
Temporal layers (TLayers) capture the temporal correlations of the same sensor across different time steps. 
For the $l$-th TLayer, the input is also the output the $(l-1)$-th STLayer, i.e., $\textbf{H}^{l-1}_{t}$, when STF takes the parallel structure. When STF takes the sequential structure, the input is the output of the $l$-th SLayer, i.e., $\textbf{H}^{sp,l}$. Similar to the SLayers, the input of node $n$ is mapped into three sub-spaces, i.e., $\textbf{Q}_{n}^{te,l}, \textbf{K}_{n}^{te,l}$, and $\textbf{V}_{n}^{te,l}$. 
Then, the attentions are computed as follows, where $\textbf{M}^{te}_{t,t'}$ is a mask to mask the messages from future time steps. The output at every time step is concatenated to form the TLayer output $\textbf{H}^{te,l}$.
{\small
\begin{equation}
\label{eq:te-sta-mask}
    \textbf{h}^{te,l}_{t,n} = \sum_{t'=1}^{T}\mathrm{softmax}(\mathrm{sim}(\textbf{Q}_{t,n}^{te,l},\textbf{K}_{t',n}^{te,l})\odot \textbf{M}^{te}_{t,t'})\textbf{V}_{t',n}^{te,l}.
\end{equation}
}

\paragraph{Spatial-temporal fusion.}
We do not need additional fusion when STF takes the sequential structure.
When STF takes the parallel structure, spatial-temporal fusion (e.g., adding, concatenating, and  gating~\cite{gatedfusion}) is applied to the outputs of the SLayer and the TLayer, to produce the output of an STLayer $\mathbf{H}^{l} = \mathrm{fuse}(\mathbf{H}^{sp,l},\mathbf{H}^{te,l}).$ 

\paragraph{Decoder.} 
After the input features are processed through $L$ (a system parameter) STLayers, the output  $\mathbf{H}^L$ of the final STLayer is fed into a decoder $g_{\omega}$,  to  forecast traffic conditions
$\hat{\textbf{X}}_{t+1:t+T'} = g_{\omega}(\textbf{H}^L)$.
In practice, the decoder can be made up of linear layers~\cite{pdformer} or self-attention layers~\cite{gman}.

Finally, a forecasting loss, e.g., the root mean squared error (RMSE), is utilised to optimise the model with Eq.~\ref{eq:pred_loss}.

\subsection{Time Complexity of \model}
\label{subsec:a_dualcast_time_cpx}
\paragraph{Time complexity.} 
\model\ takes $O(T\cdot N\cdot D^2 + \mathcal{F})$ time to map $\mathbf{X}_{t-T+1:t}$ to $\mathbf{X}_{t+1:t+T'}$, where $T$ denotes the length of the input (output) time window, $N$ denotes the number of sensors, $D$ is the feature dimensionality of the hidden layers, and $\mathcal{F}$ denotes the time complexity of the spatial-temporal models (GMAN~\cite{gman}, STNN~\cite{sttn}, or PDFormer~\cite{pdformer}). The first term $T\cdot N\cdot D^2$  accounts for the cost of linear projections applied to all input matrix.
Note that $D$ is usually smaller than $N$ in traffic forecasting tasks. Given that the time complexity of the spatial-temporal models is typically $O(\mathcal{F})=O(T\cdot N^2 \cdot D)$, the time complexity of \model\ then becomes $O(T\cdot N^2 \cdot D)$. This time complexity suggests that applying \model\ to power existing spatial-temporal models will not increase their time complexity, although there is a hidden constant of 2, i.e., \model\ has two branches of spatial-temporal models. 

\subsection{Rooted Sub-tree Cross-time Attention}
\label{subsec:a_rct}

\begin{figure}[!t]
    \centering
    \begin{subfigure}[b]{0.3\linewidth}
        \includegraphics[width=\linewidth]{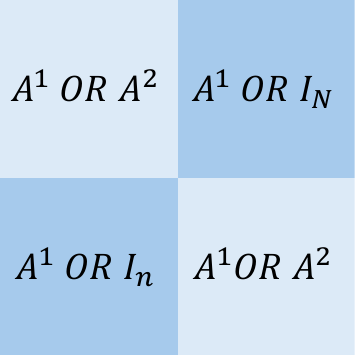}
    \caption{}
    \end{subfigure}
    \hspace{-0.05mm}
    \begin{subfigure}[b]{0.3\linewidth}
        \includegraphics[width=\linewidth]{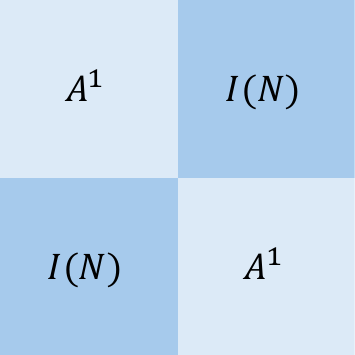}
    \caption{}
    \end{subfigure}
    \caption{ Adjacency matrices for different model structures}
    \label{fig:ct-adj}
\end{figure}

\paragraph{Adjacency matrices used for local attention computation.} As Fig.~\ref{fig:ct-adj} shows, there are two ways to build the adjacent matrices $\mathbf{M}$ (not a matrix of 1's any more) used in Eq.~\ref{eq:sp-sta-fm} for local attention computation in \model.
In the figure, $\mathbf{A}^1$ is the original adjacency matrix that comes with the input graph $G$ (Section~\ref{sec:Preliminaries}) and $\mathbf{A}^k$ the $k$-th-order matrix of $\mathbf{A}^1$ indicating the connectivity between $k$-hop neighbours.

Fig~\ref{fig:ct-adj}(a) elaborates a simple way to learn high-order relationships, i.e., to pre-compute all $k$-hop adjacency matrices to connect all neighbours for up to $k$ hops. However, this approach ignores the local hierarchical information. For example, the red dashed line and the green dotted line have different traffic propagation patterns and propagation time costs in Fig.~\ref{fig:ct_att}(a), as the red dashed line only concerns the same node across different times, while the green dotted line concerns nodes at different space and times, which should not be ignored. This approach cannot such a scenario as all neighbours are considered to share the same impact, regardless of their distances (i.e., number of hops). Our cross-time attention uses the adjacency matrix shown in Fig.~\ref{fig:ct-adj}(b) combined with $k$-step message passing to distinguish nodes at different hop distances, as discussed in Section~\ref{subsec:ct-attention}

\section{Additional Details on the Experiments}
\subsection{Detailed Experiment Setup}
\label{subsec:a_exp_setup}

\paragraph{Datasets.}
We use two freeway traffic datasets and an urban traffic dataset: 
\textbf{PEMS03} and \textbf{PEMS08}~\cite{pems} contain traffic flow data collected by 358 and 170 sensors on freeways in North Central and San Bernardino (California) from September to November 2018 and from July to August 2016, respectively;  \textbf{Melbourne}~\cite{STSM} contains traffic flow data collected by 182 sensors in the City of Melbourne, Australia, from July to September 2022.

The traffic records in PEMS03 and PEMS08 are given in 5-minute intervals, i.e., 288 intervals per day, while those in Melbourne  are given in 15-minute intervals, i.e., 96 records per day. {Table~\ref{tab:dataset_info}} summarises the dataset statistics. As shown, Melbourne has a higher standard deviation and is more challenging dataset.

\begin{table}[!t]
\small
\centering
\begin{tabular}{l|c|c|c}
  \hlineB{3}
  Dataset & PEMS03 & PEMS08 & Melbourne\\
  \hline \hline
  \#sensors & 358 & 170 & 182\\
  \#edges & 547 & 277 & 398\\
  Mean of readings & 179.26 & 230.68 & 115.74 \\
  Std Dev. of readings & 143.71 & 146.2 & 155.86\\
  Median of readings & 136 & 215 & 56\\
  Starting time & 09/2018 & 07/2016 & 07/2022\\
  End time & 11/2018 & 08/2016 & 09/2022\\
  Time interval & 5 min & 5 min & 15 min\\
  \hlineB{3}
\end{tabular}
\caption{Dataset Statistics}
\label{tab:dataset_info}
\end{table}

\paragraph{Implementation details.} We use the default settings from the source code for both the baseline models and their variants powered by \model. 
We optimise the models by using the Adam optimiser with a learning rate starting at 0.001, each in 100 epochs. 
For the models powered by \model, we use grid search with a staged strategy on the validation sets to determine the hyper-parameters values $\alpha$, $\beta$, and $\gamma$. The search range $\{0.01, 0.05, 0.1, 0.5, 1, 5\}$ follows previous works~\cite{transgtr,liu2022contrastive}.
Table~\ref{tab:parameters} lists the tuned values of the hyper-parameters used in experiments.

\begin{table}[!t]
\small
\centering
\begin{tabular}{p{1.5cm}|c|c|c|c}
  \hlineB{3}
  Model & Parameter & PEMS03 & PEMS08 & Melbourne\\
  \hline \hline
  \multirow{3}{*}{STTN} 
  & $\alpha$ & 0.01 & 1 & 0.01\\
  & $\beta$ & 0.01 & 0.01 & 0.01\\
  & $\gamma$ & 0.5 & 0.05 & 0.1\\
    \hline
  \multirow{3}{*}{GMAN} 
  & $\alpha$ & 1 & 5 & 0.05\\
  & $\beta$ & 0.1 & 0.1 & 0.1\\
  & $\gamma$ & 0.5 & 0.1 & 5\\
  \hline
  \multirow{3}{*}{PDformer} 
  & $\alpha$ & 0.05 & 0.1 & 0.05\\
  & $\beta$ & 0.01 & 0.1 & 0.01\\
  & $\gamma$ & 0.1 & 1 & 0.1\\
  \hlineB{3}
\end{tabular}
\caption{Parameter Settings}
\label{tab:parameters}
\end{table}

\subsection{Overall results}
\label{subsec:a_over_results}
\begin{figure}[!t]
    \centering
    \begin{subfigure}[b]{0.24\textwidth}
        \includegraphics[width=\textwidth]{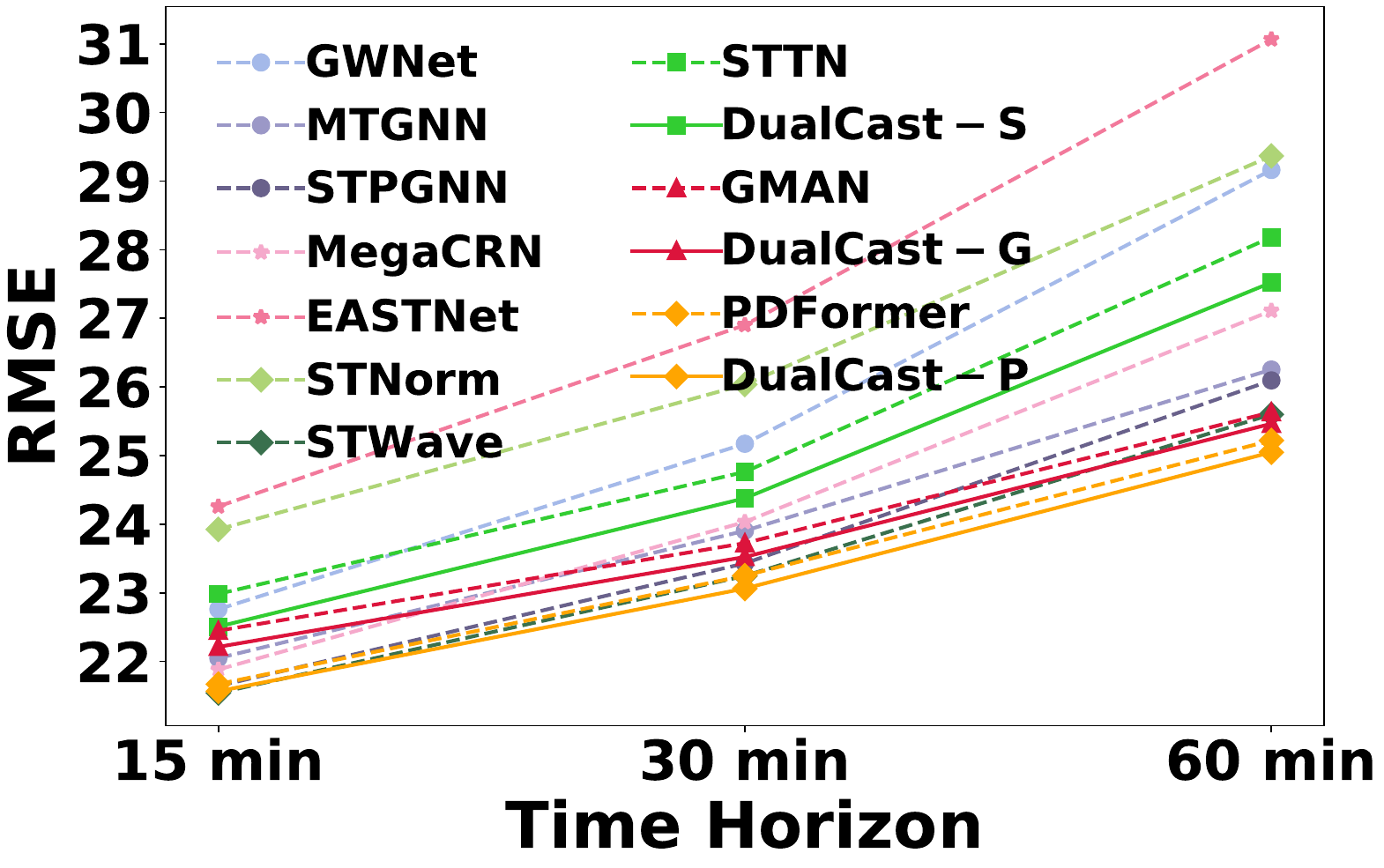}
    \caption{PEMS08}
    \end{subfigure}
\hspace{-0.13cm}
    \begin{subfigure}[b]{0.24\textwidth}
        \includegraphics[width=\textwidth]{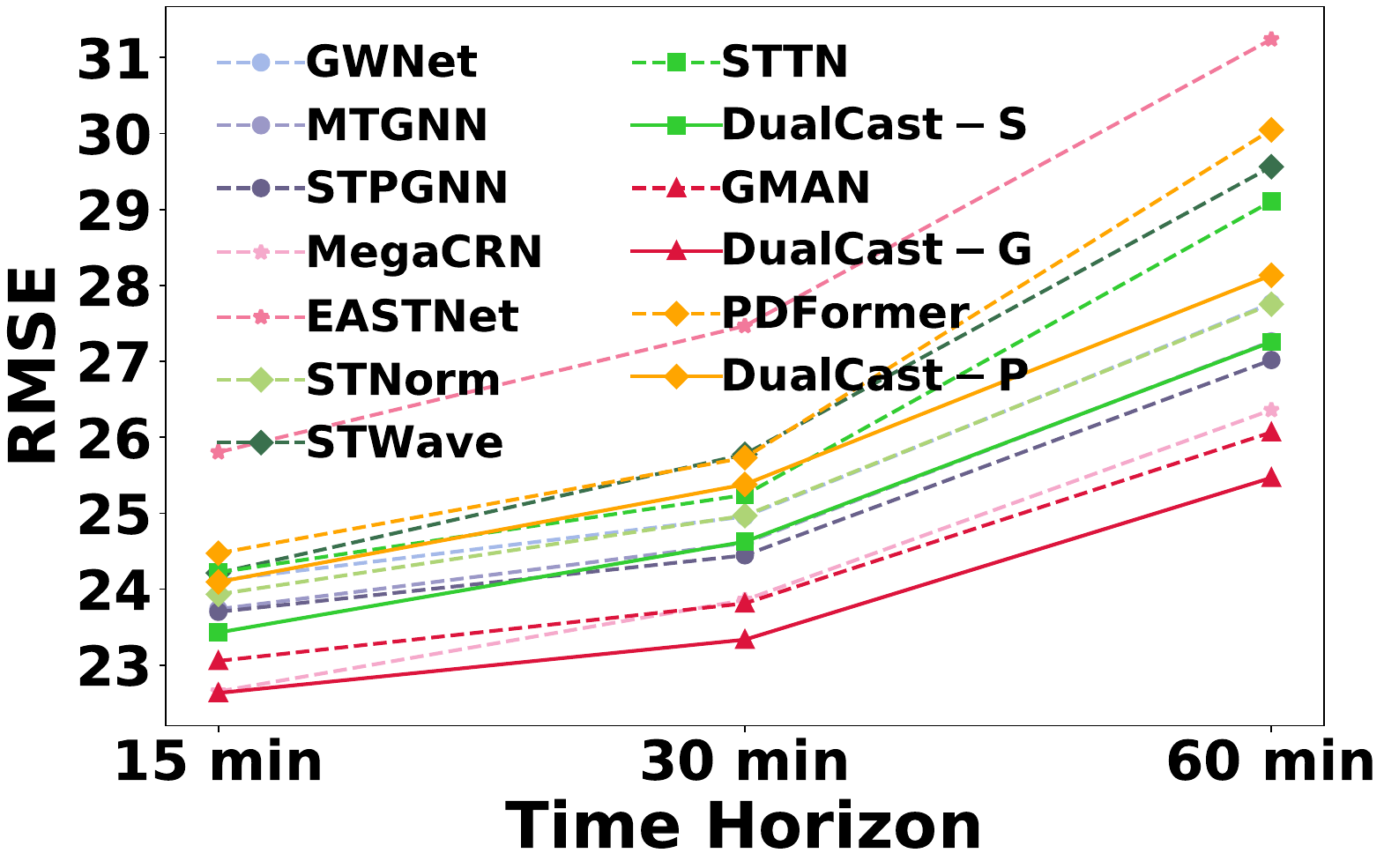}
    \caption{Melbourne}
    \end{subfigure}
    \caption{\small Forecasting errors vs. time horizon}
    \label{fig:rmse_horizon}
\end{figure}

\paragraph{Model performance across all times.} Fig.~\ref{fig:rmse_horizon} shows the forecasting errors of different time horizons on PEMS08 and Melbourne. The results show that the models powered by \model\ outperform the corresponding original models consistently, while \model-P is the most effective model on PEMS08 and \model-G is most effective model on Melbourne dataset. We see that \model\ outperform the baseline models consistently at different forecasting horizons, confirming their effectiveness. 

\paragraph{Model scalability.}
We vary sensor numbers from 200 to 1,000 in  Melbourne, expanding coverage from downtown to suburbs. We use Melbourne as PEMS03 and PEMS08 have bounded areas with limited sensors.

\begin{figure}[t!]
         \centering
  \includegraphics[width=0.5\columnwidth]{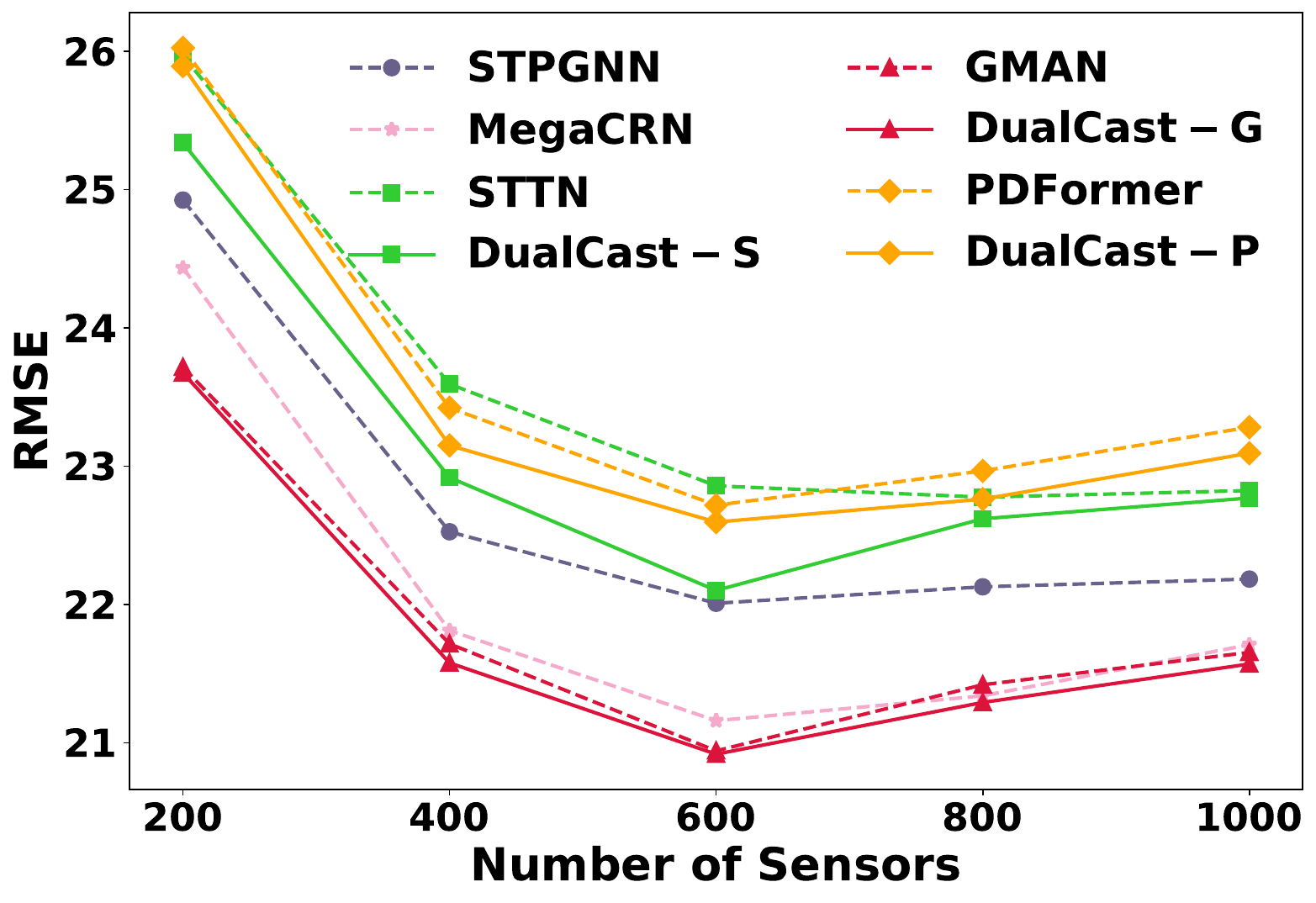}
         \caption{Forecasting errors vs. number of sensors}
         \label{fig:vary_sensor_number}
\end{figure}

Fig.~\ref{fig:vary_sensor_number} shows the forecasting errors. Like before, the model variants powered by \model\ outperform the vanilla counterparts consistently, while the best \model-based model, \model-G, also outperforms the best baseline models.
When the number of sensors grows, the errors first drop and then increase. The initial drop is because we start with sensors in the City of Melbourne, where there is a high level of variation in the traffic data leading to high forecasting errors. These high errors get amortised as more sensors from outer suburbs are added. As more and more sensors are added (e.g., over 600), the learning capabilities of the models get challenged, which causes the errors to rise again. 

\paragraph{Effectiveness vs. model scale.} 
\model\ has a dual-branch structure that increases the number of model parameters. 
We conduct further experiments to compare the model effectiveness between the \model-based models and the original models without using \model\ but with 
twice or triple the number of layers (such that the numbers of parameters of each pair of models differ by less than 6\%), to show that \model\ is more effective than simply doubling the size of the models. Table~\ref{tab:param_vs_permformance} shows the results on PEMS03, where \textbf{STTN-D}, \textbf{GMAN-D}, and \textbf{PDFormer-D} denotes the model variants with twice the number of parameters of STTN, GMAN, and PDFormer, respectively. We see that these model variants are outperformed by our \model-based variants in RMSE, even though both types of variants share similar numbers of parameters.
These results show that simply doubling the size of the models is not as effective as our dual-branch model design, thus emphasising the advantage of our model structure. We omit the performance results on the other two datasets as the comparative model performance patterns are similar. 
\begin{table}[!t]
\small
\centering
\resizebox{\columnwidth}{!}{
\begin{tabular}{l|c|c|l|>{\columncolor{gray!20}}c}
\hlineB{3}
\textbf{Model} & \textbf{\#Parameter} &\textbf{\#Training Time (s)} & \textbf{RMSE} & \textbf{Error reduction} \\ \hline
\hline
STTN-D         & 540,581 & 210.1 & 29.960        &                                              \\
DualCast-S    & 558,252  & 271.3  &26.282 & \multirow{-2}{*}{12.3\%}                        \\ \hline
GMAN-D        &  569,667 & 618.3& 26.590  &                                              \\
DualCast-G   & 535,804    & 756.5 & 25.582  & \multirow{-2}{*}{3.8\%}                        \\ \hline
PDFormer-D     & 1,114,077   & 426.8 & 26.180   &                                              \\
DualCast-P   & 1,174,918  & 730.8  & 24.898     & \multirow{-2}{*}{4.9\%}                        \\ \hlineB{3}
\end{tabular}
}
\caption{Model Effectiveness vs. Scale (PEMS03)}
\label{tab:param_vs_permformance}
\end{table}

\begin{figure}[!t]
    \centering
    \begin{subfigure}[b]{0.23\textwidth}
        \includegraphics[width=\textwidth]{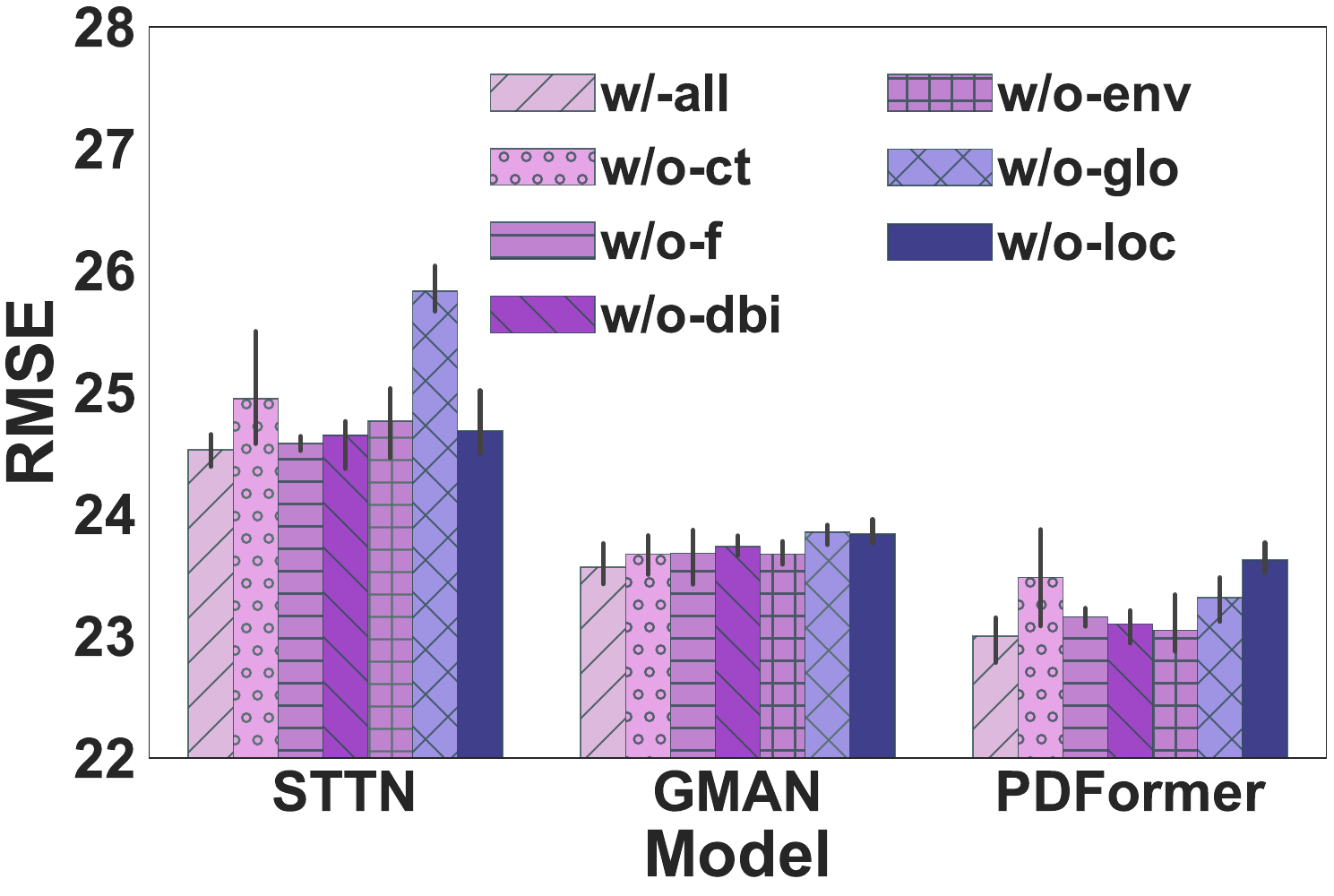}
    \caption{PEMS08}
    \end{subfigure}
\hspace{-0.1cm}
    \begin{subfigure}[b]{0.23\textwidth}
        \includegraphics[width=\textwidth]{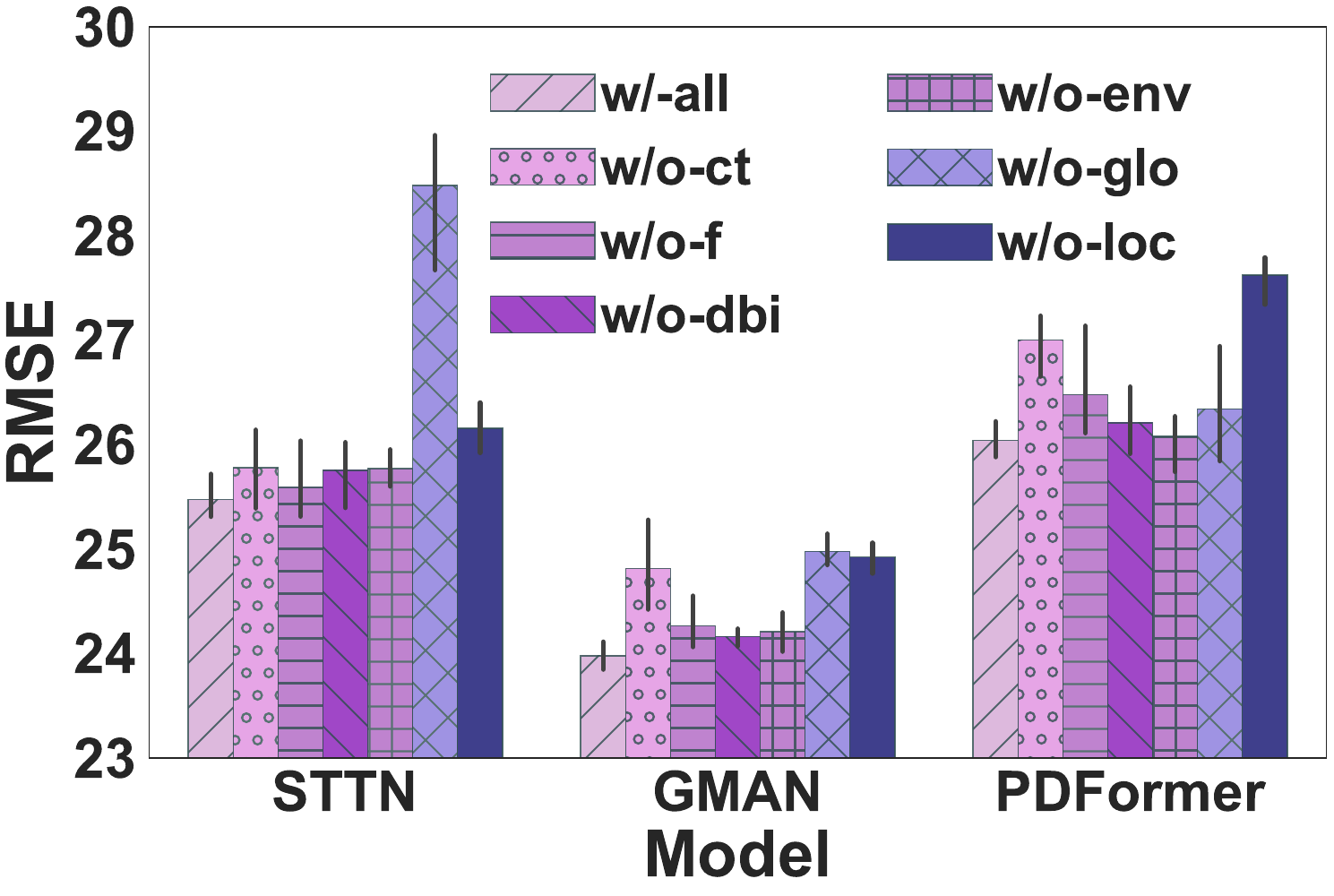}
    \caption{Melbourne}
    \end{subfigure}
    \caption{\small Ablation Study}
    \label{fig:ablation_study}
\end{figure}

\subsection{Ablation Study}
\label{subsec:a_ablation_study}
\paragraph{Effectiveness of all components.}
Fig~\ref{fig:ablation_study} presents ablation study results on PEMS08 and Melbourne. All modules in \model\ contribute to the overall model performance on this dataset, again confirming the effectiveness of the model components. 
Cross-time attention is more important for GMAN and PDFormer on Melbourne due to its high fluctuations (Table~\ref{tab:dataset_info}), as it helps capture high-order spatial-temporal relationships for faster responses to those sudden changes. For STTN, which already employed GCN and self-attention for spatial correlations, cross-time attention is less important. Similarly, the GCN-based spatial feature extraction reduces the need for local attention in \model-S.
However, local attention is crucial for \model-P because PDFormer employs heterogeneous masking-based attention, which effectively captures diverse local spatial information. Without it, hierarchical road network details and semantic correlations are lost, significantly degrading performance. 

\begin{figure}[t]
    \centering
    \begin{subfigure}[b]{0.98\columnwidth}
        \includegraphics[width=0.98\columnwidth]{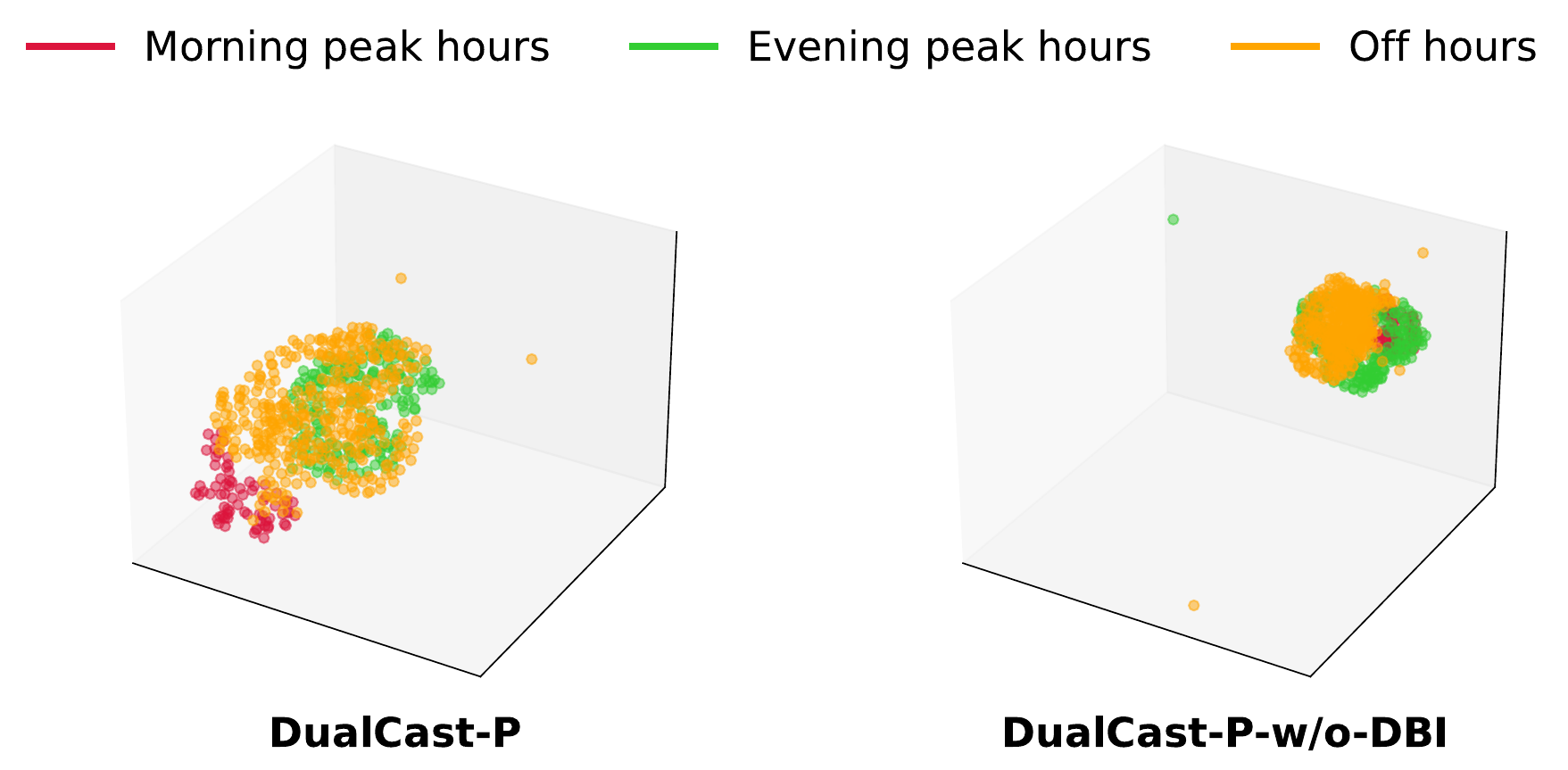}
    \caption{Comparing representations on Monday}
    \end{subfigure}

    \begin{subfigure}[b]{0.98\columnwidth}
        \includegraphics[width=0.98\columnwidth]{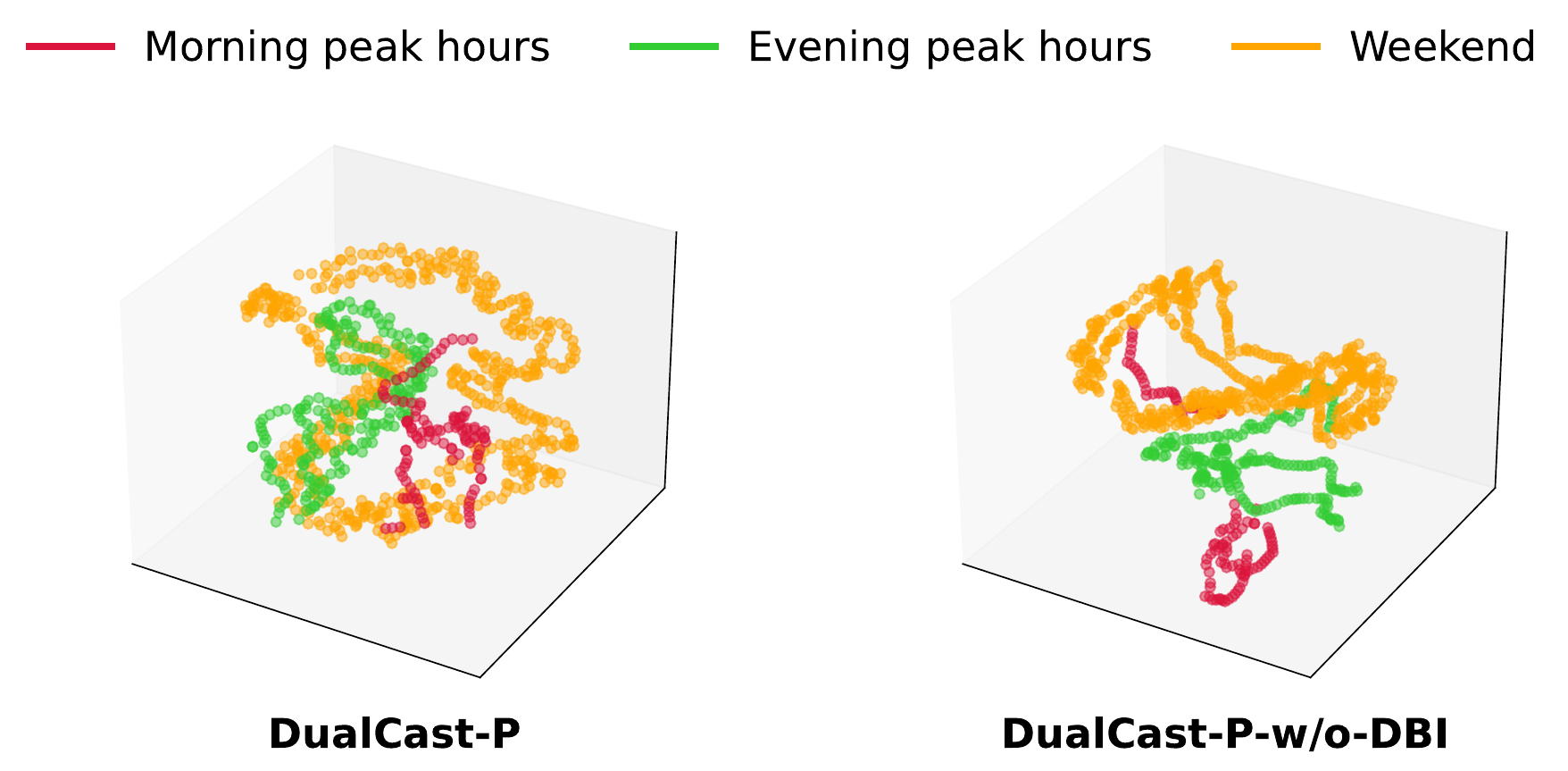}
    \caption{Comparing representations on Friday and Saturday}
    \end{subfigure}
\caption{t-SNE visualisation on the feature space for \model-P and \model-P-w/o-DBI on PEMS03.}
\label{fig:dbi_effectiveness}
\end{figure}

\paragraph{Effectiveness of representation learning for different patterns.} 
Fig.~\ref{fig:dbi_effectiveness}(a) visualises the representations learned by \model-P with and without the DBI loss on PEMS03 during the morning peak hour (06:00 to 09:00, in red), evening peak hour (16:00 to 22:00, in green) and off-hour (in orange) on all Mondays in the test set (same as below), to verify if the DBI loss can guide the model to distinguish traffic patterns within the same day. Fig.~\ref{fig:dbi_effectiveness}(b) further visualises representations during the morning peak hour (in red), evening peak hour (in green) on all Fridays and representations during Saturdays (in orange), to verify if the DBI loss can guide the model to distinguish traffic patterns between workdays and weekends. As the figures show, \model-P with the DBI loss generates representations for different times and days that are well separated, while \model-P-w/o-DBI mixes these representations. This result confirms the effectiveness of the DBI loss. We omit results on the other two datasets as their performance patterns are similar.

\begin{figure}[!t]
    \centering
        \includegraphics[width=0.48\textwidth]{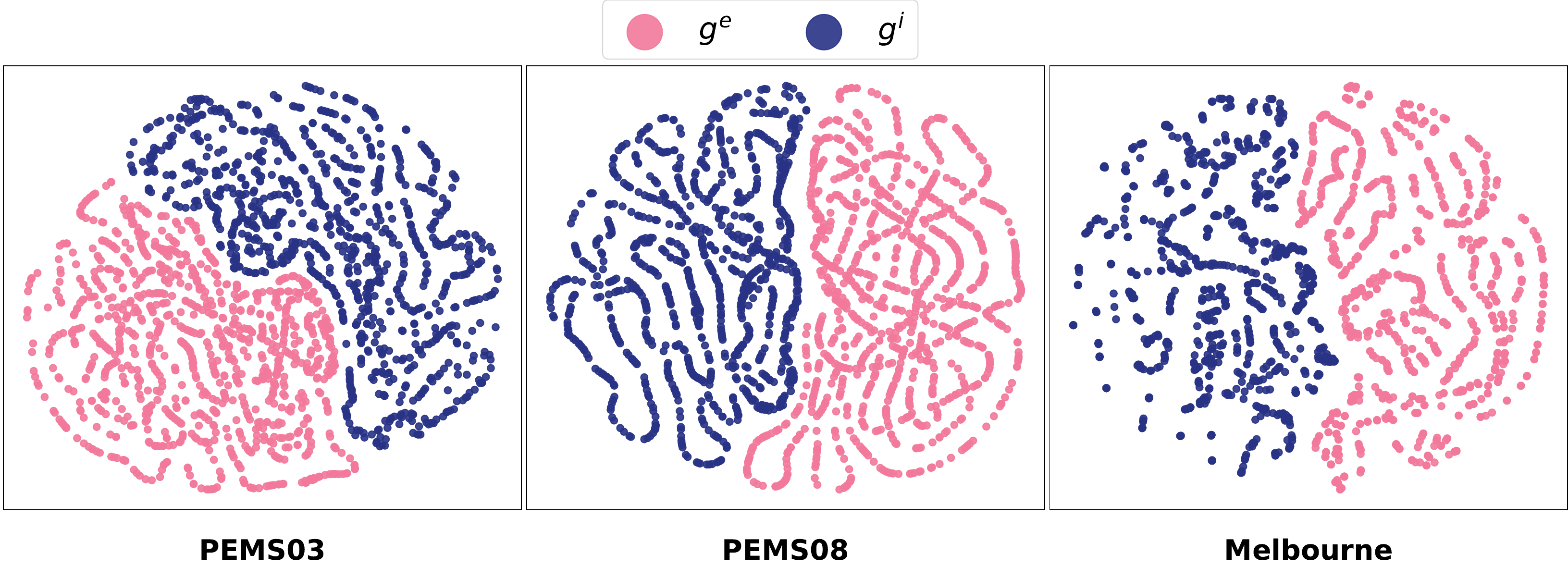}
    \caption{t-SNE visualisation for $\textbf{g}^{i}$ and $\textbf{g}^{e}$ of \model-P.}
    \label{fig:disentangle_ge_gi}
\end{figure}

\paragraph{Effectiveness of disentanglement.} To verify the effectiveness of disentanglement with our dual-branch structure, we adopt t-SNE to reduce the dimension of $\textbf{g}^{i}$ and  $\textbf{g}^{e}$ produced by \model-P (recall that these are the outputs of the two branches) to 2 and visualise them in Fig.~\ref{fig:disentangle_ge_gi}. 
The blue dots in the figure represent  $\textbf{g}^{i}$, and the pink dots represent $\textbf{g}^{e}$. As the figure shows, our proposed disentanglement can separate the intrinsic signals from the environmental context, further confirming the model's effectiveness.

\begin{figure}[!t]
    \centering
    \begin{subfigure}[b]{0.45\textwidth}
        \includegraphics[width=\textwidth]{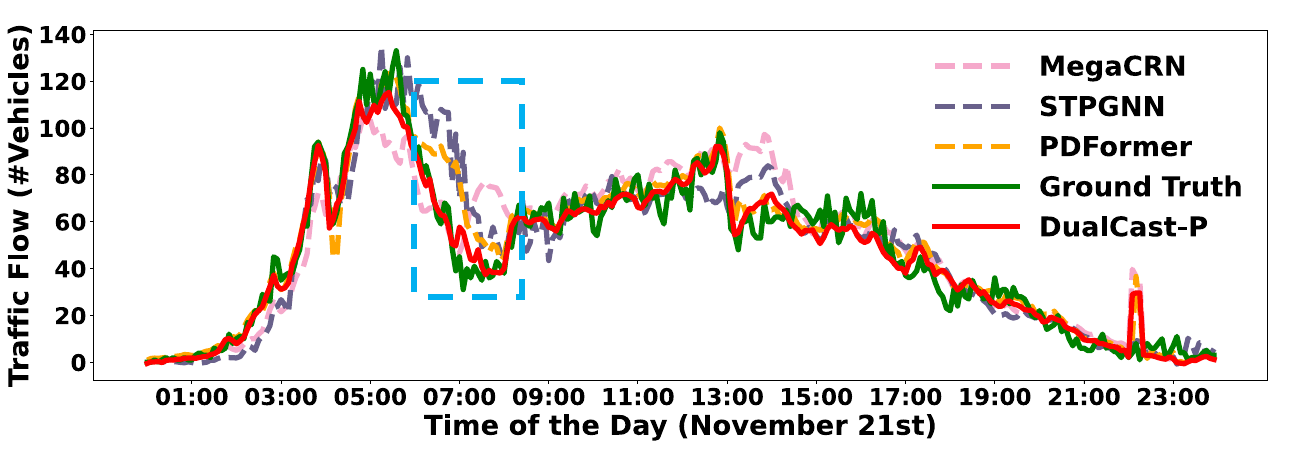}
    \end{subfigure}
    \caption{A case study of responding to sudden changes (highlighted in rectangles) in traffic at sensor \#72 on PEMS03 on Nov. 21.}
    \label{fig:a_sudden_change}
\end{figure}

\begin{figure}[!t]
    \centering
    \begin{subfigure}[b]{0.32\columnwidth}
        \includegraphics[width=\columnwidth]{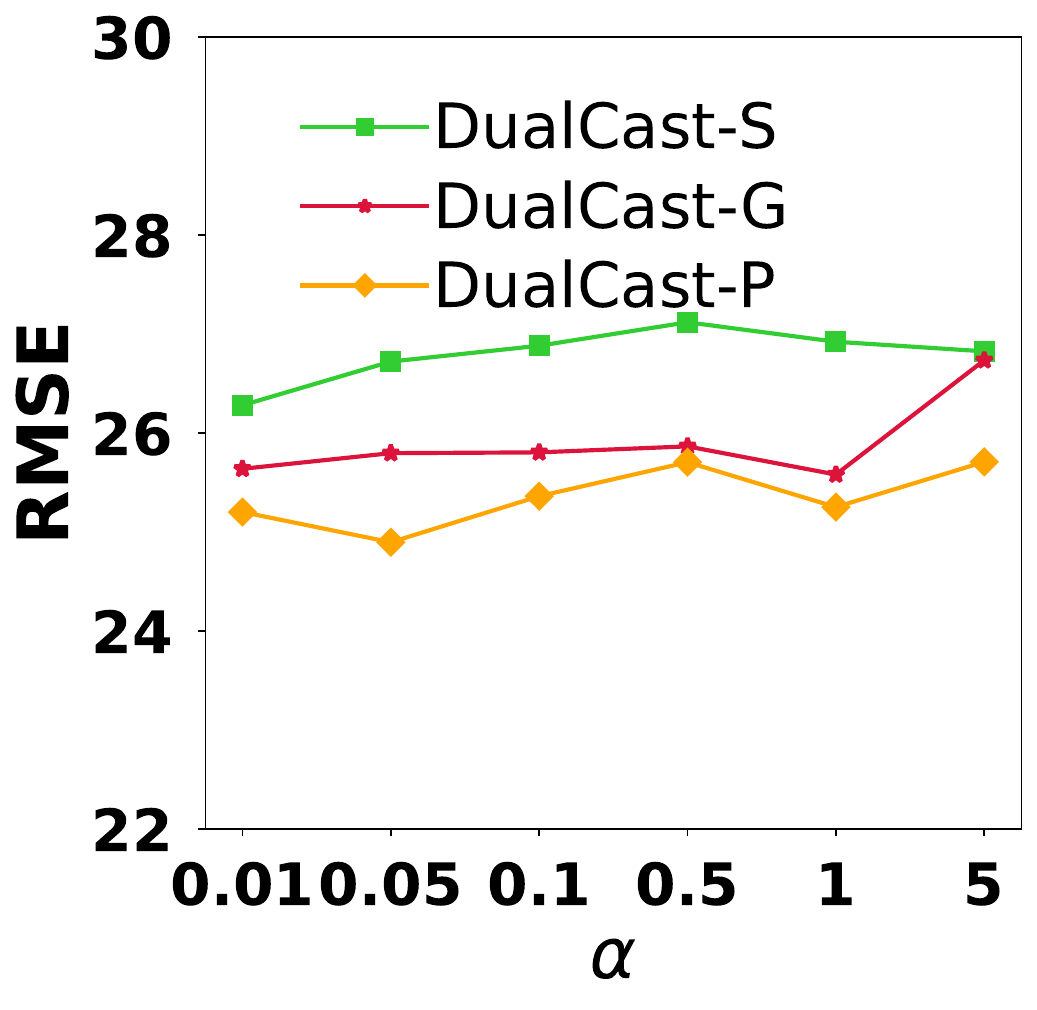}
    \caption{PEMS03}
    \end{subfigure}
    \begin{subfigure}[b]{0.32\columnwidth}
        \includegraphics[width=\columnwidth]{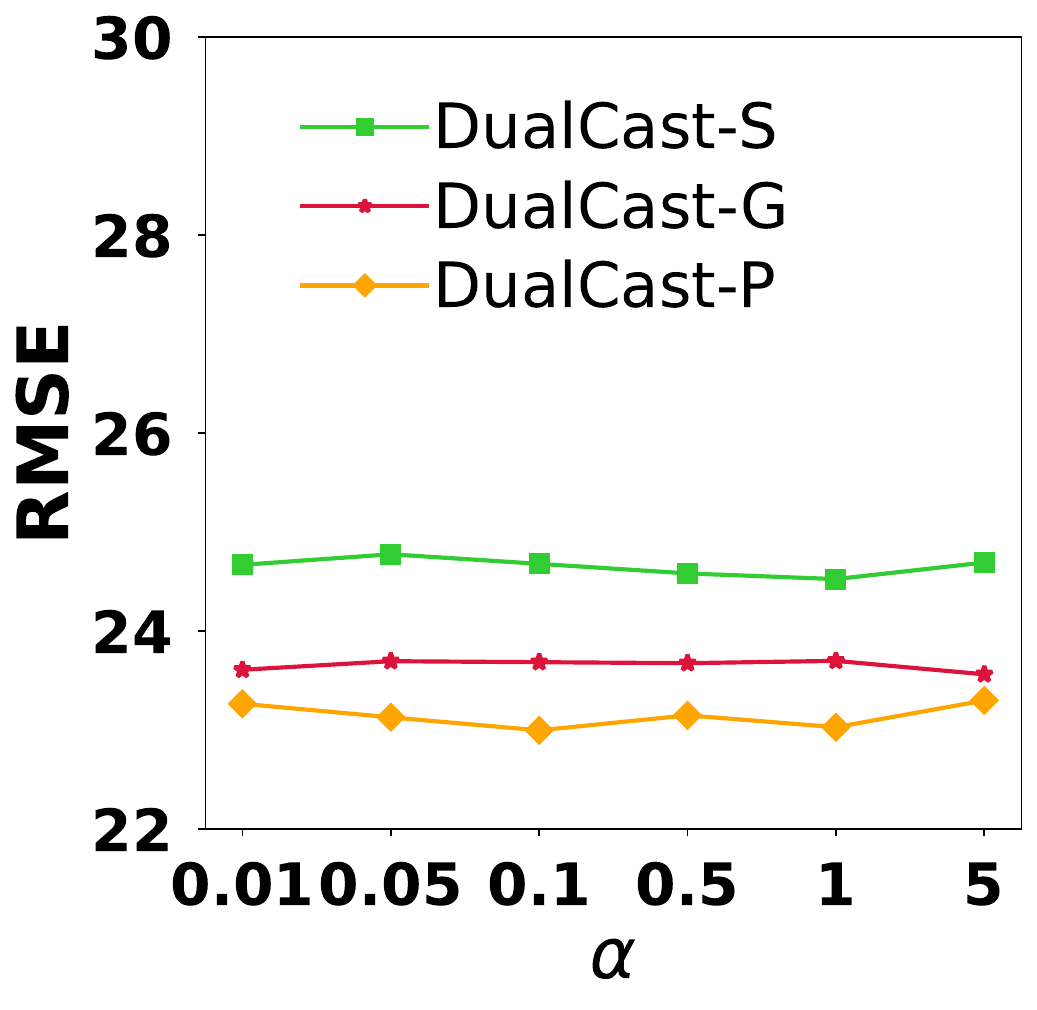}
    \caption{PEMS08}
    \end{subfigure}
    \begin{subfigure}[b]{0.32\columnwidth}
        \includegraphics[width=\columnwidth]{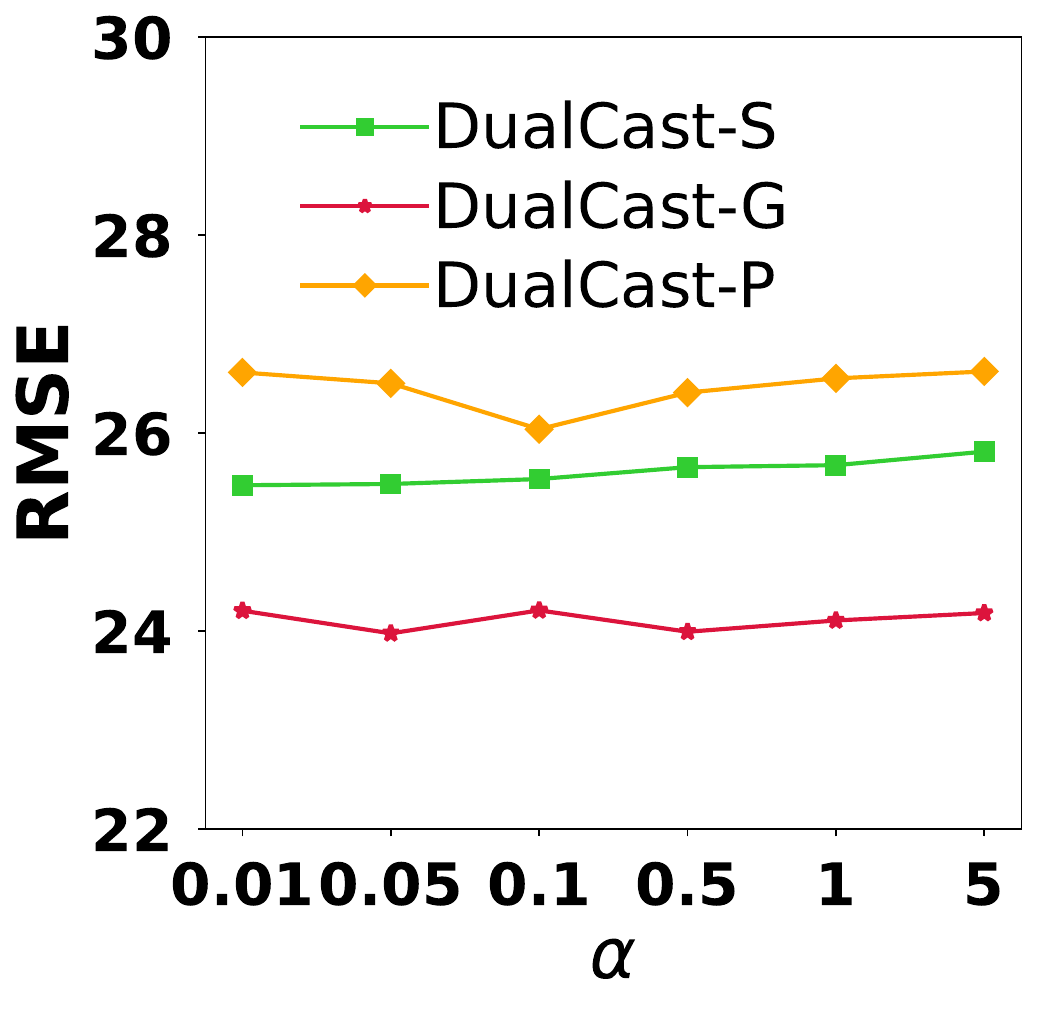}
    \caption{Melbourne}
    \end{subfigure}
    \caption{Forecasting errors vs. hyper-parameter $\alpha$ in the loss}
    \label{fig:param_study_alpha}
\end{figure}

\begin{figure}[!t]
    \centering
    \begin{subfigure}[b]{0.32\columnwidth}
        \includegraphics[width=\columnwidth]{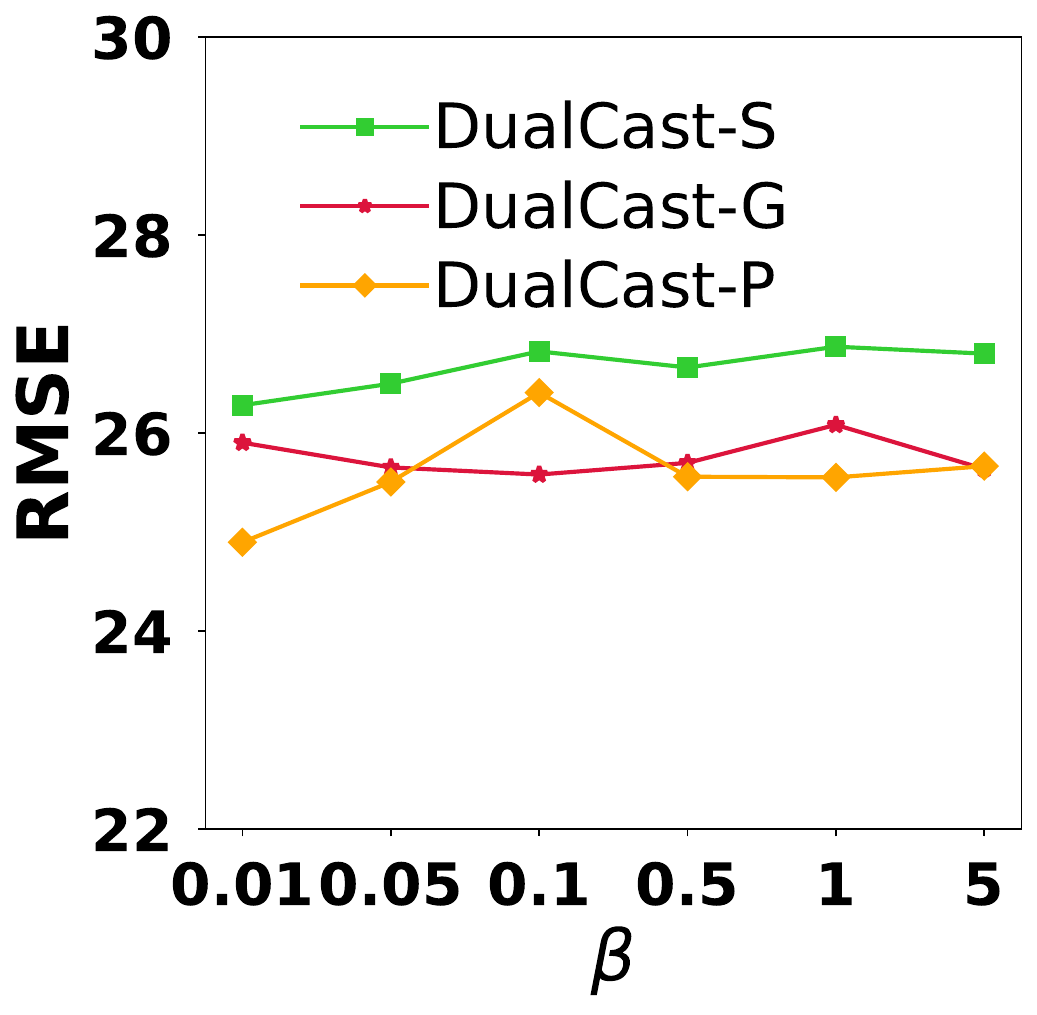}
    \caption{PEMS03}
    \end{subfigure}
    \begin{subfigure}[b]{0.32\columnwidth}
        \includegraphics[width=\columnwidth]{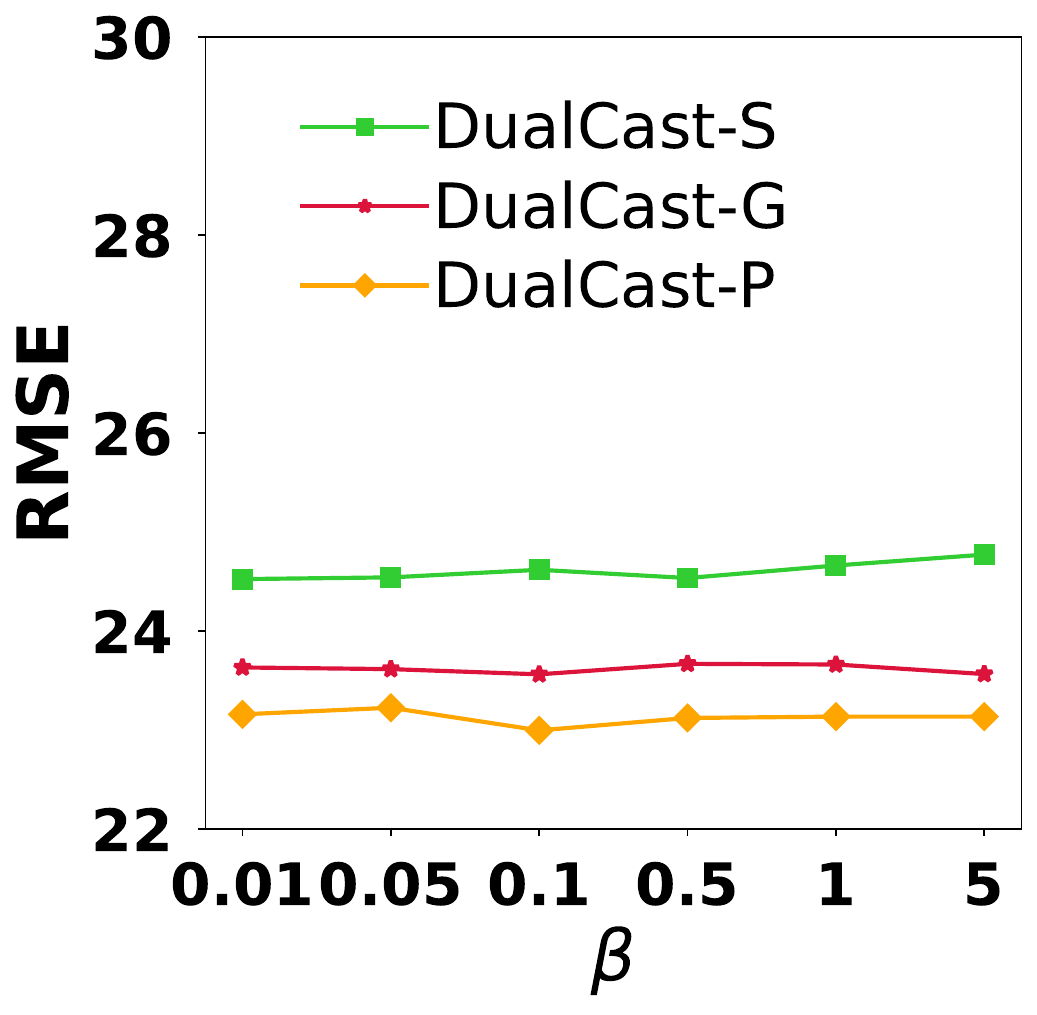}
    \caption{PEMS08}
    \end{subfigure}
    \begin{subfigure}[b]{0.32\columnwidth}
        \includegraphics[width=\columnwidth]{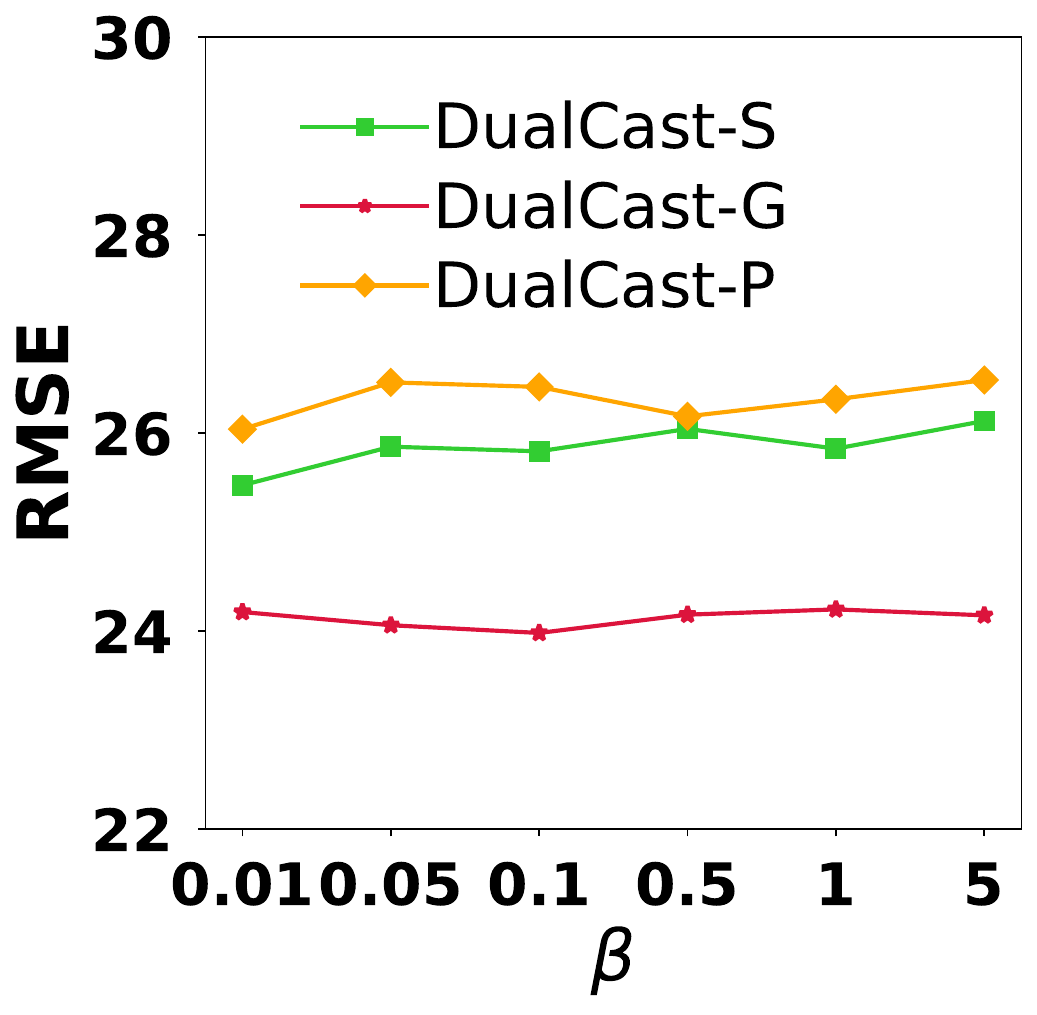}
    \caption{Melbourne}
    \end{subfigure}
    \caption{Forecasting errors vs. hyper-parameter $\beta$ in the loss}
    \label{fig:param_study_beta}
\end{figure}

\begin{figure}[!t]
    \centering
    \begin{subfigure}[b]{0.32\columnwidth}
        \includegraphics[width=\columnwidth]{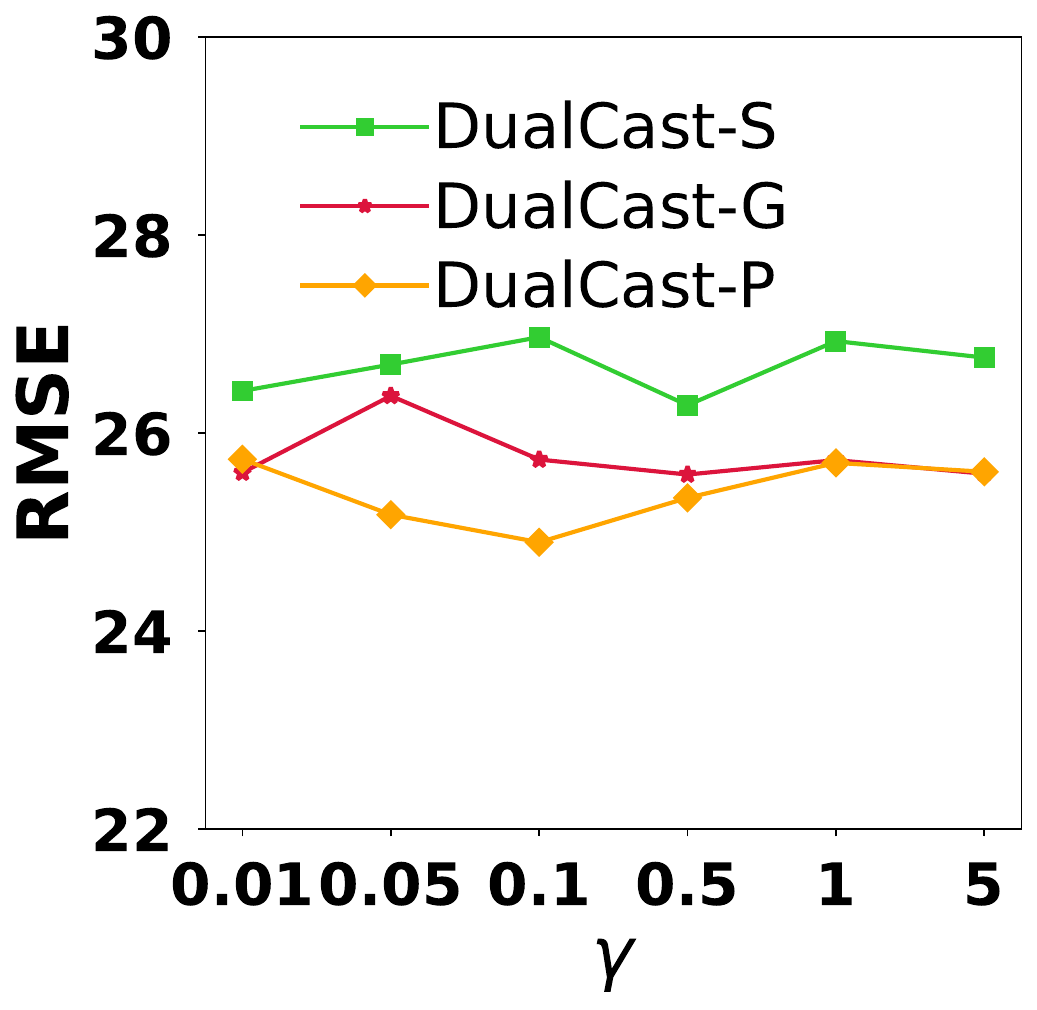}
    \caption{PEMS03}
    \end{subfigure}
    \begin{subfigure}[b]{0.32\columnwidth}
        \includegraphics[width=\columnwidth]{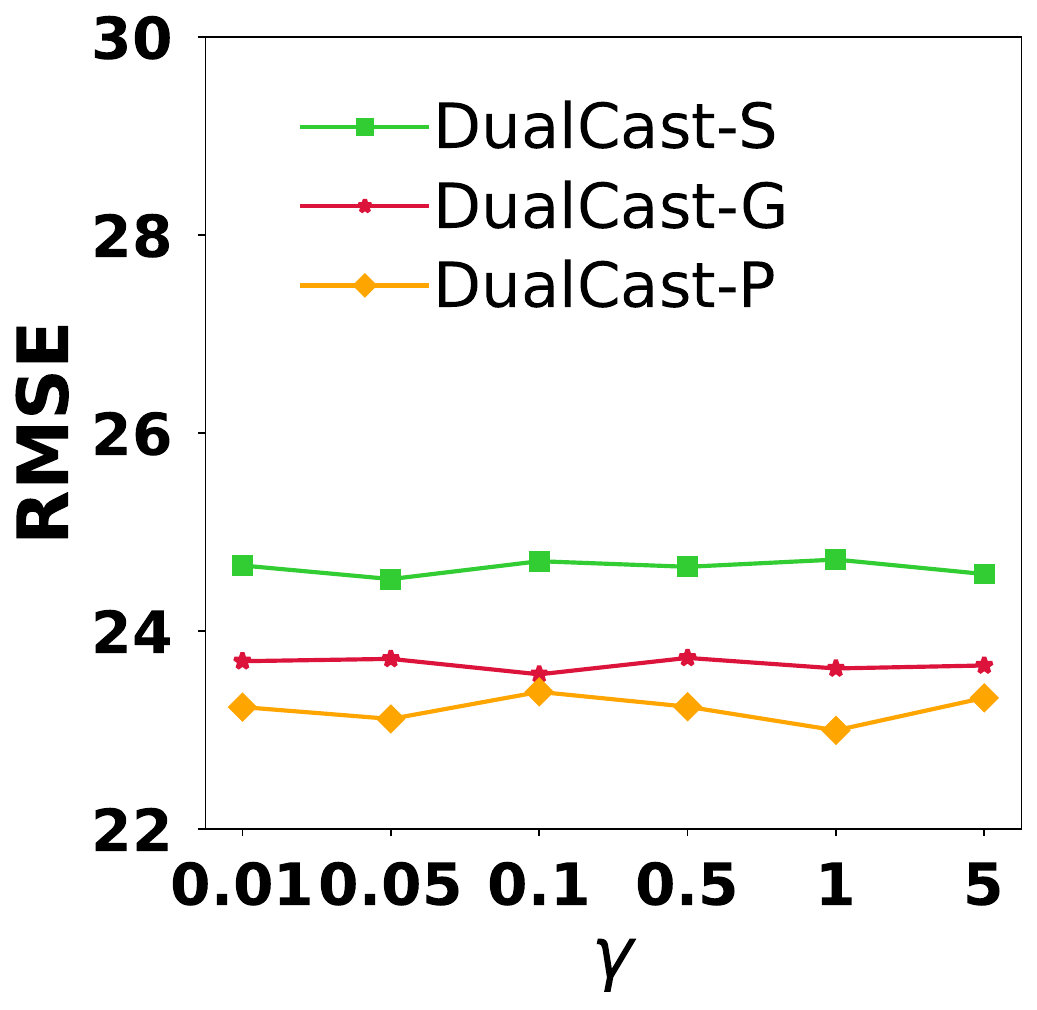}
    \caption{PEMS08}
    \end{subfigure}
    \begin{subfigure}[b]{0.32\columnwidth}
        \includegraphics[width=\columnwidth]{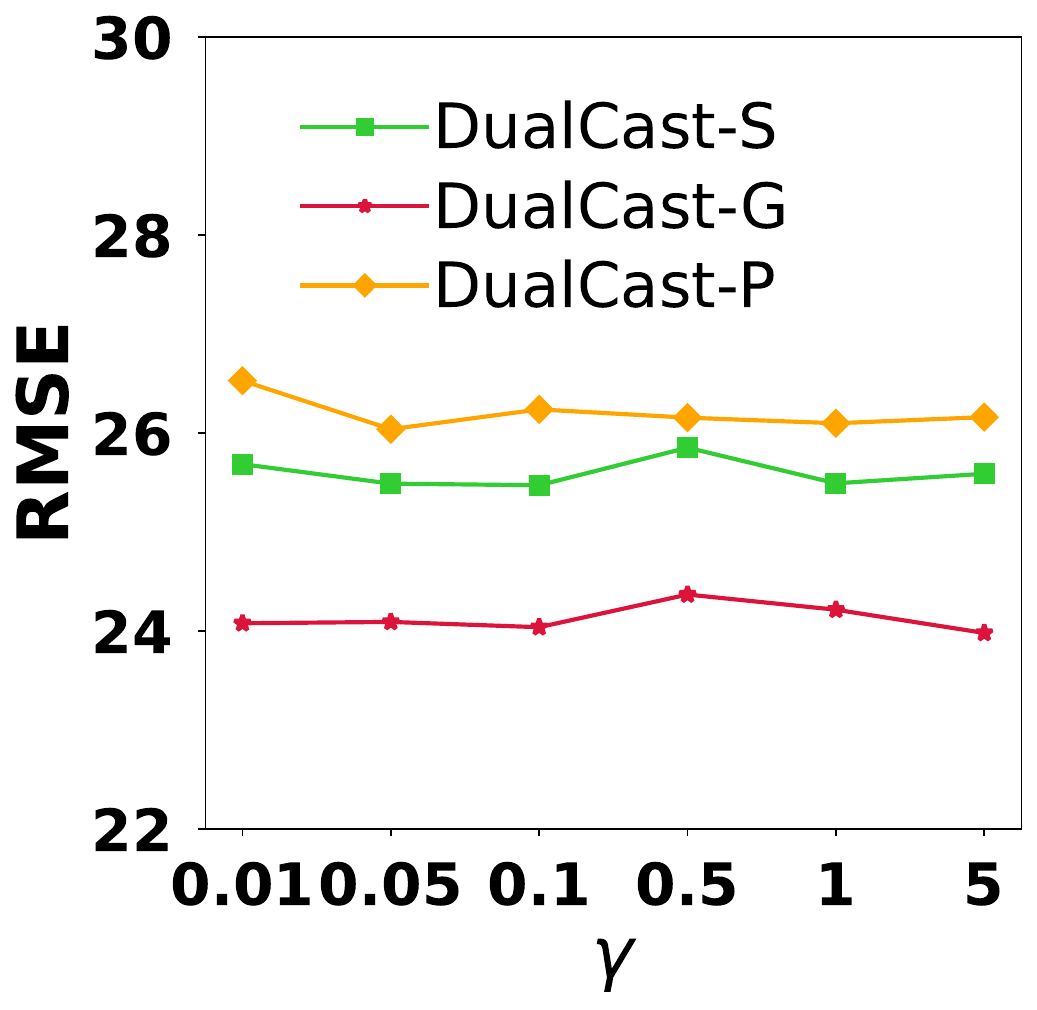}
    \caption{Melbourne}
    \end{subfigure}
    \caption{Forecasting errors vs. hyper-parameter $\gamma$ in the loss}
    \label{fig:param_study_gamma}
\end{figure}

\begin{table}[!t]
\centering
\setlength\columnsep{0pt}
\resizebox{\columnwidth}{!}{
\begin{tabular}{l|l|l|ccc}
\hlineB{3}
\multirow{2}{*}{\textbf{Dataset}} & \multicolumn{1}{c|}{\multirow{2}{*}{\textbf{Model}}} & \multicolumn{1}{c|}{\multirow{2}{*}{\textbf{RMSE}}} & \multicolumn{3}{c}{\textbf{Time Cost (s)}}                     \\ \cline{4-6} 
                                  & \multicolumn{1}{c|}{}                                & \multicolumn{1}{c|}{}                               & \textbf{Training} & \textbf{Inference} & \textbf{Spatial Att.} \\ \hline\hline
\multirow{2}{*}{PEMS03}           & \model-G-sim                                                  & 26.510                                              & 1467.9            & 28.0                 & 1093.8                \\
                                  & \model-G-rct                                                  & 25.900                                              & 412.8             & 21.5               & 76.6                  \\ \hline
\multirow{2}{*}{PEMS08}           & \model-G-sim                                                  & 23.727                                              & 194.3             & 5.9                & 86.1                  \\
                                  & \model-G-rct                                                  & 23.683                                              & 125.8             & 5.4                & 22.3                  \\ \hline
\multirow{2}{*}{Melbourne}        & \model-G-sim                                                  & 24.320                                              & 35.8              & 1.3                & 16.9                  \\
                                  &\model-G-rct                                                  & 24.183                                              & 23.3              & 0.9                & 6.3                   \\ \hlineB{3}
\end{tabular}
}
\caption{Spatial Attention Efficiency (\model-G)}
\label{tab:att_time}
\end{table}

\paragraph{Efficiency of cross-time attention.}
Next, we test the attention computation efficiency. We use \model-G as GMAN's own attention module is the simplest compared with PDFormer and STTN, making it more suitable to observe the impact of the cross-time attention added by \model.
We implement two variants as described in Section~\ref{subsec:ct-attention}: 
(i)~\textbf{\model-G-sim} computes the spatial self-attention with the adjacency matrix shown in Fig.~\ref{fig:ct-adj}(a). This adjacency matrix models the relationships between neighbours within two hops at the same time step, as well as relationships between neighbours within one hop at different time steps.
(ii)~\textbf{\model-G-rct} (this is the one using our proposed rooted sub-tree attention) computes the attention and updates the representation with message-passing. This model uses the adjacency matrix shown in Fig.~\ref{fig:ct-adj}(b), which models the relationships between neighbours within one hop at the same time step and the relationships between a node and itself at different time steps. 

Table~\ref{tab:att_time} reports the results.  \model-G-sim takes more time for model training, inference, and attention computation, yet it has higher errors compared with \model-G-rct. These results confirm that our proposed rooted sub-tree attention is more efficient and effective for computing the high-order spatial-temporal correlations and the cross-time attention.

\subsection{Parameter and Case Study}
\label{subsec:a_param_study}
\paragraph{Parameter study.}
We study the impact of the hyper-parameters in our final loss function (Eq.~\ref{eq:loss_final}), namely $\alpha$, $\beta$, and $\gamma$. Figs.~\ref{fig:param_study_alpha} to~\ref{fig:param_study_gamma} show that, overall, the forecasting errors stay at low values consistently when the hyper-parameters in the loss vary. Such results confirm that the three losses jointly contribute to the model accuracy and that \model\ is robust without the need for heavy hyper-parameter tuning.

\paragraph{Responding to sudden changes in traffic.}
On the PEMS datasets, ground-truth traffic events are unavailable. Instead, we inspected the ground-truth traffic observations and found two representative sensors ($\#72$ and $\#97$) on PEMS03 with sudden changes on November 21st, 2018.
Fig.~\ref{fig:a_sudden_change} shows the ground-truth traffic flow and 1-hour-ahead forecasts by MegaCRN, STPGNN, PDFormer, and \model-P at Sensor $\#72$. \model-P provides the closest forecasts to the ground truth, especially during sudden traffic changes, again highlighting the strength of \model.

\begin{table}[!t]
\small
\centering
\begin{tabular}{l l |ccc }
\hlineB{3}
\multicolumn{1}{l|}{{\textbf{Dataset}}}         & {\textbf{Method}} & {\textbf{Overall}} & {\textbf{60 min}} & {\textbf{cpx}} \\
\hline
\hline
\multicolumn{1}{l|}{{}}                         & {STTN}            & {27.170}               & {30.386}              & {40.990}            \\
\multicolumn{1}{l|}{\multirow{-2}{*}{{PEMS03}}} & {STTN-dual}          & {26.950}               & {29.872}              & {39.350}            \\ \hline
\rowcolor{gray!20}
\multicolumn{2}{l|}{Error reduction}                                                                          & {0.8\%}                  & {1.7\%}                 & {4.0\%}               \\ \hline\hline
\multicolumn{1}{l|}{{}}                         & {STTN}            & {24.980}               & {28.182}              & {31.400}            \\
\multicolumn{1}{l|}{\multirow{-2}{*}{{PEMS08}}} & {STTN-dual}          & {24.900}               & {28.090}              & {30.820}            \\ \hline
\rowcolor{gray!20}
\multicolumn{2}{l|}{Error reduction}                                                                          & {0.3\%}                  & {0.3\%}                 & {1.8\%}               \\ \hline\hline
\multicolumn{1}{l|}{{}}                         & {STTN}            & {26.400}               & {29.108}              & {30.550}            \\
\multicolumn{1}{l|}{\multirow{-2}{*}{{Melbourne}}}    & {STTN-dual}          & {25.770}               & {28.360}              & {29.000}            \\ \hline
\rowcolor{gray!20}
\multicolumn{2}{l|}{Error reduction}                                                                          & {2.4\%}                  & {2.6\%}                 & {5.1\%}               \\ 
\hlineB{3}
\end{tabular}
\caption{Effectiveness (RMSE) of the dual-branch structure on STTN. ``Overall'' refers to errors averaged over 1 hour, ``60 min'' to errors at the 1-hour horizon for all-day windows. ``cpx'' to errors during complex times (4:00 pm to 8:00 pm).}
\label{tab:sttn_gcn}
\end{table}

\paragraph{Effectiveness of our dual-branch structure when applying to GCN-based models.} We further explore the application of \model\ to GCN-based models. We use STTN for this set of experiments since it utilises GCN for spatial feature extraction. Here, we apply only the dual-branch structure and its corresponding loss functions to STTN, without incorporating the cross-attention mechanism. We denote this variant as \textbf{STTN-dual}. As Table~\ref{tab:sttn_gcn} shows, STTN-dual consistently outperforms STTN on all datasets. Importantly, STTN-dual also achieves larger improvements at complex times, which again confirms that our proposed dual-branch structure can effectively handle complex situations (aperiodic events).

\end{document}